# LEAK DETECTION IN NATURAL GAS PIPELINES USING INTELLIGENT MODELS.

BY

OSHINGBESAN ADEBAYO
MATRIC NUMBER: 179579

A Project in the Department of Petroleum Engineering,
Submitted to the Faculty of Technology
In partial fulfilment of the requirements for the Degree of
BACHELOR OF SCIENCE

Of the
UNIVERSITY OF IBADAN

APRIL 2019

# CERTIFICATION

This is to certify that Mr. Oshingbesan Adebayo with matriculation number 179579 of the Department of Petroleum Engineering, Faculty of Technology, University of Ibadan, carried out under our supervision.

……………………………………........                                              …………………….

**Project Supervisor**                                                                                    Date

**Dr. O. Akinsete,**

B.Sc. (Ibadan), M.Sc., Ph.D. (Ibadan)

Department of Petroleum Engineering,

University of Ibadan, Nigeria.

……………………………………........                                              …………………….

**Head of Department**                                                                                    Date

**Prof. S. O. Isehunwa,**

B.Sc. (Ibadan), M.Sc., Ph.D. (Ibadan)

Department of Petroleum Engineering,

University of Ibadan, Nigeria.

# DEDICATION

I'd like to dedicate this work to God, family and everyone who has been a part of my growth journey in the University of Ibadan.

# ACKNOWLEDGEMENTS

I want to acknowledge everyone who has been a part of my growth story in the University of Ibadan starting from the Lord God Almighty, in whom we breathe, in whom we live and in whom we have our existence

I would like to specially thank my supervisor, Dr Akinsete for his guidance and help throughout the period of this project. I also want to thank all the wonderful lectures and laboratory personnel of the department of Petroleum Engineering for their impact in my life.

I will also say a big thank you to Ibadan Varsity Christian Union for their love and care. To every of my family and friends, I duly appreciate your care and love. Space will fail me to mention names. I love you all.

Lastly, I also want to acknowledge the Class of 2018 – Perspe Nation for the wonderful four/five years we have spent together. The journey would have been less fun than it was without you guys.

Thank you everyone.

# ABSTRACT

Leak detection in gas pipelines is an important and persistent problem in the Oil and Gas industry. This is particularly important as pipelines are the most common way of transporting natural gas. The Oil and Gas industry is beginning to investigate how tools of Data Science, Machine Learning/ Intelligent Models, Artificial Intelligence, Big Data, Cloud Computing etc. can be used to improve current industry processes.

The project aims to study the ability of data-driven intelligent models to detect small leaks for a natural gas pipeline using basic operational parameters such as pressure, temperature and flowrate and then compare the intelligent models among themselves using existing performance metrics (sensitivity, reliability, robustness and accuracy). This project applies the observer design technique to detect leaks in natural gas pipelines using a regresso-classification hierarchical model where an intelligent model acts as a regressor and a modified logistic regression acts as a classifier. Five intelligent models (gradient boosting, decision trees, random forest, support vector machine and artificial neural network) are studied in this project using a pipeline data stream of four weeks.

The results shows that while support vector machine and artificial neural networks are better regressors than the others, they do not provide the best results in leak detection due to their internal complexities and the volume of data used. The random forest and decision tree models are the most sensitive as they can detect a leak of 0.1% of nominal flow in about 2 hours. All the intelligent models had high reliability with zero false alarm rate in testing phase. However, due to this, the models had low accuracy with the artificial neural network and support vector machine having the best accuracy. All the intelligent models have fairly good robustness. The average time to leak detection for all the intelligent models was compared to a real time transient model in literature. The intelligent models performs comparatively well when all trade-offs are taken into account.

The results show that intelligent models performs relatively well in the problem of leak detection. This research is useful to academicians, engineers etc. who are looking at ways of improving leak detection results while harnessing the tools of big data, data analytics, artificial intelligence etc. This result suggests that intelligent models could be used alongside a real time transient model to significantly improve leak detection results.

# Table of Contents

# LIST OF FIGURES

# LIST OF TABLES

# NOMENCLATURE

| | |
|---|---|
| A | Area of the Pipeline (sqft) |
| c | Sonic Velocity (ft/s) |
| D | Diameter of Pipe (in) |
| $f$ | Friction Factor |
| g | Acceleration due to gravity (ft/s$^2$) |
| h | Enthalpy (J/kg) |
| L | Pipe Length (miles) |
| m | Mass rate (lb/s) |
| M | Molecular Mass (lbmol) |
| P | Pressure (psia) |
| R | Gas Constant (ft3.psi.R-1lbmol-1) |
| T | Temperature |
| $u$ | Velocity (ft/s) |
| z | Compressibility factor |

Greek Symbol

| | |
|---|---|
| $\rho g$ | Fluid density of gas (lb/ft3) |
| $\mu$ | Gas viscosity (cp) |
| $\gamma g$ | Gas specific gravity |

# CHAPTER ONE

## INTRODUCTION

### 1.1 Overview

Over the ages, transportation of goods has been one of human's basic needs. For the transportation of fluids, pipelines have proven since the 400 BC to be a suitable method. For natural gas, pipelines are still the most efficient and cost-effective way of transporting medium to large volumes over short to medium distances. They are virtually everywhere. While they are mostly buried, their flexibility in terms of the range of terrains (under the sea, through a desert, across a swamp etc.) makes them the most common means of transporting natural gas. They are also perhaps the safest means of transporting natural gas.

However, over time, unintended harms can come from the use of pipelines due to leaks caused by corrosion, environment, external parties etc. Many of these harms may be severe in nature affecting the environment, damaging properties, causing injuries and maybe even loss of lives. With about three million kilometres of pipelines running constantly worldwide, leak detection in pipelines is key to minimize the effect of these harms. Work is continually ongoing to improve the accuracy of leak detection and location in order to allow for a swift and efficient response. The efficiency of these systems can be assessed based on its accuracy, reliability, robustness, and sensitivity.

The Oil and Gas industry is said to be one of the largest generators of data in terms of volume after the likes of Google, Facebook, and Amazon etc. While these companies have found ways of learning from these data and making better and smarter decisions, the Oil and Gas industry still lags much more behind. However, the industry is beginning to realize the need for adaptation as major players are beginning to form Big Data partnerships to see how their volume of data can yield better-informed decisions. As shown by the biggest technology companies in the world, data in the hands of intelligent models can work wonders in terms of improving the effectiveness and efficiency of processes.

## 1.2 Natural Gas

Natural gas has been used commercially as a fuel for centuries in China. It became one of the leading fuels all over the world over the last century. Natural gas is composed majorly of methane, ethane and less often, propane and butane alongside some very minute quantities of heavier hydrocarbon components. There may also be non-hydrocarbon constituents such as nitrogen, hydrogen sulphide, carbon dioxide, water vapour and so on. Though it is usually buried thousands of feet below ground, engineers have developed techniques of producing them to the surface either solely or with oil and to deliver it in the best possible state to the market.

## 1.3 Pipelines

Pipelines are closed systems that transport fluids commodities from one location in space and time to another. It includes all physical devices, components, computer systems, telecommunication systems and the pipe itself (Harrie et al., 2016). Fundamentally, pipelines are simple. They connect a place of higher pressure to another of lower pressure. However, they can be added complexities. Equipment like pumps, compressors may be used to provide additional pressure increase. Tanks may provide temporary storage (even the pipe could act as a tank of sort), valves may be used to divert flow, prevent backflow and topology/ terrain may differ greatly.

Generally, there are three types of pipelines:

1. <u>Gathering lines</u>: They usually consist of low pressure, small pipelines that transport the raw natural gas from the wellhead to the processing plants.

2. <u>Transmission lines</u>: They usually consist of high pressure, large pipelines that transport natural gas from the processing plants to the centres of consumption.

3. <u>Distribution lines</u>: They are similar to gathering lines. They deliver gas to the final consumer.

### 1.31 Pipeline Pack

Apart from being a transportation means, the pipeline can also act as a storage facility. This is particularly important in gas pipelines because of their high compressibility. This phenomenon is called pipeline pack. There is no way to physically measure the pipeline pack thus is usually assumed constant or approximated using all available pressure and temperature measurements in combination with an equation of state for the fluid.

## 1.4 Data Acquisition and Control

While physical components of a pipeline are important, they are of little or no use if they cannot be monitored, operated and controlled. To achieve this, one would require field monitoring equipment such as pressure, temperature and flow meters to provide us with information concerning the pipeline operating conditions. Usually, a pipeline is monitored and controlled from a single site using a control environment. The control environment consists of the human controller, the Supervisory Control and Data Acquisition (SCADA) system and one or more remote site data controllers such as Remote Terminal Units (RTU), Programmable Logic Controllers (PLC) (Figure 1.2).

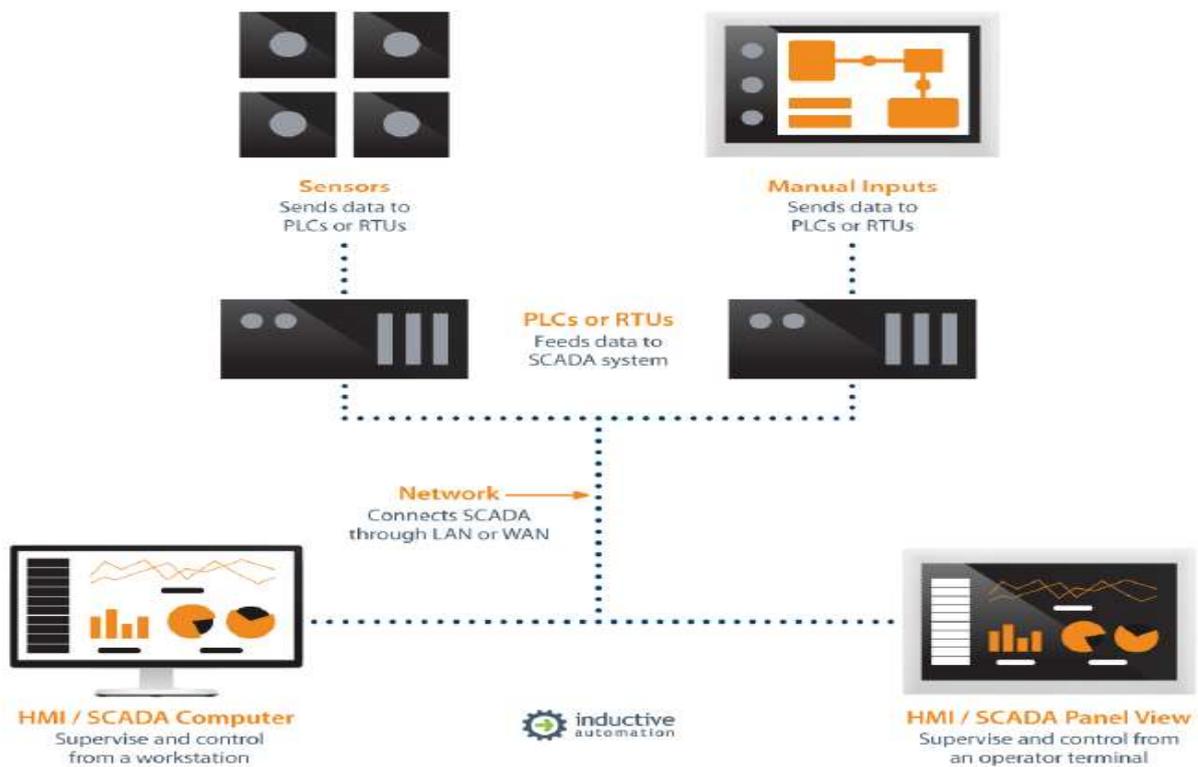

Figure 1.1 SCADA Architecture (Inductive Automation)

## 1.5 Leakages

Leak events are generally very rare but do occur and can cause major damages. A leak might be small and gradual or large and sudden. While there may be many causes of leaks, there are four major categories of causes:

1. *Pipeline corrosion and wear*

   Fatigue cracks are one cause. These occur as the result of material fatigue and are often found on longitudinal welds. Tensile strength can cause stress tears which can reduce the effectiveness of cathodic corrosion protection systems, resulting in corrosion on the pipeline. Stress corrosion is another possible cause. Cracks can also be caused by hydrogen indexing.

2. *Operation outside design limit*

   Failure to operate within design specifications has also been known to lead to pipeline leakages. Operations at excessive pressure, flowing corrosive fluids, etc. are examples of failure to adhere to operation guidelines.

3. *Sabotage (Intentional damage)*

   The world's oil and gas pipelines pass within and across regions that are poor, politically unstable, inhospitable, corrupt, or prone to terrorist attacks.

4. *Unintentional third-party damage*

   Leaks can also occur when an external force acts from the outside. This is the case when backhoes dig up a pipeline or seismic ground movements cause shifts in the ground surrounding a pipeline.

## 1.6 Leak Detection Approaches

There are a wide range of approaches to leak detection, each having its own strengths, weaknesses, and costs. The major divisions are:

1) Physical Observation
2) Mass Balance Systems
3) Real-Time Transient Models
4) Rarefaction Wave Models
5) External Leak Detection Systems (e.g. Fibre-Optic Cables)
6) Intermittent Leak Detection Systems (e.g. Intelligent Pigs)
7) Adjunct Systems (e.g. Pressure Point Analysis, Fuzzy systems)

## 1.7 Big Data, Data Science and Intelligent Models

Big data refers to any collection of data sets so large or complex that it is cannot be processed by using traditional data management techniques such as relational database management systems. Data science is the use of mathematical, statistical, computing and every other related tool to analyse massive amounts of data and extract the knowledge that it contains in order to make smart and informed decisions.

Three important characteristics of big data are

1) Volume – how much data is there?
2) Variety – How diverse is the data there?
3) Volume – at what speed is new data generated?

The data science workflow consists majorly of six things:

1) Research Objective
2) Retrieving Data
3) Preparing Data
4) Exploring Data
5) Modelling Data
6) Presenting Information from Data and Automation

Intelligent models are algorithms designed to learn from large volumes of data and draw valuable insights from them. Examples are neural networks, support vector machines, neuro-fuzzy systems, decision trees, decision forests, recommender systems etc. They are termed intelligent because like the human brain, there is only an intuitive understanding of how and why they can learn from data.

Generally, they are classified into

1) <u>Regression models</u>: These are models that predict a real-valued output given a real-valued input. They could be linear or non-linear. A common application is predicting demand for a product.
2) <u>Clustering models</u>: These are models that group data into different groups or clusters. They are usually used for market segmentation problems.
3) <u>Classification models</u>: these are models that output a discrete value output given a real-valued input. They can be used for handwriting recognition.

## 1.8 Problem Statement

Leak detection within a gas pipeline is difficult to simulate because of the behaviour of the properties of a gas. Continuous work has been ongoing as regards the improvement of leak detection models. With the boom of intelligent models and the volume of data generated by a pipeline system, there is still room for improvement in the sensitivity, accuracy, robustness, and reliability of leak detection systems.

## 1.9 Aims and Objectives

This project seeks to examine the application of intelligent models in the field of leak detection in natural gas pipelines.

The specific objectives include:

1) To study the ability of intelligent models to detect small leaks for a gas pipeline using basic operational parameters (e.g. pressure, temperature, and flow rate).
2) To compare the intelligent models among themselves and an available literature data using existing industry performance metrics.

## 1.10 Justification for the Study

Organizations like Google, Amazon, and Facebook etc. have shown that historical data is a goldmine that needs to be used efficiently and effectively. As the Oil and Gas industry begins to move away from its more conservative roots to the much more dynamic world of data, petroleum engineers should begin to study how tools of Data Science, Big Data, Cloud Computing, Machine learning/ Intelligent models, Artificial Intelligence, Virtual Reality and so on can be used to solve existing and recurrent problems in the industry. One of such problems is leak detection in gas pipelines. Intelligent models have been used to great effect in similar problems to leak detection such as corrosion prediction is pipelines, gas demand in distribution lines and leak detection in oil pipelines. Many natural gas pipelines around the world have operating data for as much as three decades. The use of these volumes of data alongside intelligent models and other aforementioned tools should improve our ability to quickly detect and locate small leaks in most efficient and effective way possible.

# CHAPTER TWO

## LITERATURE REVIEW

### 2.1 LEAK DETECTION SYSTEMS (LDS).

Leak detection systems have a wide definition. While some argue that it consists of all the architecture that makes up a pipeline system (figure 2.1), some others argue that it should consist only of the component of the architecture (whether external systems, the pipeline, communicator channel, the data aggregator or the data processing system) that actually does the leak detection. It is worth noting that companies that offer leak detection solutions only offer the component of the architecture that performs the actual leak detection. Yet, no leak detection system is complete (or work optimally) without the entire architecture being in place and working effectively and efficiently.

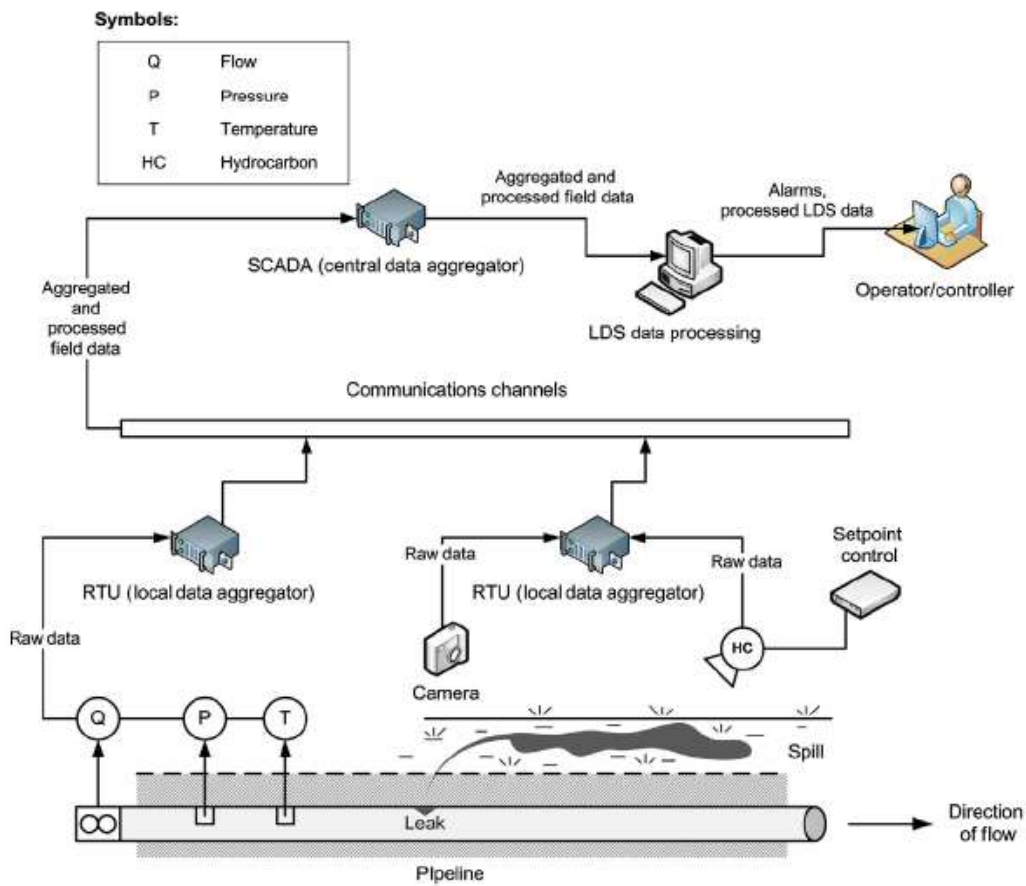

Figure 2.1 A Leak Detection Architecture (Morgan et al., 2016)

### 2.1.1 FUNCTIONS OF A LEAK DETECTION SYSTEM

Fundamentally, any LDS is only required to detect a valid leak and not create a false alarm.

However, it is often required to:

  i)   Estimate the leak rate,
  ii)  Estimate leak location,

Some other functions can also be imposed on it. These may include but not limited to:

  i)    Assisting in diagnosing the alternate cause of the leak alarm
  ii)   Calculating an urgency-oriented metric
  iii)  Offering an alternate causal hypothesis
  iv)   Calculation of process measurements in locations they are not available
  v)    Tracking of pipeline commodities and pig

### 2.1.2 Leak Detection Methods

Based on the components in the leak detection architecture that performs leak detection, leak detection can be classified into different methods. Generally, they are classified into external/ direct methods and internal/inferential methods. Specifically, they can be classified as:

1) *Hardware-Based Systems*

This refers to the all the models they are driven by hardware devices to detect and localize a leak. Examples include radiotracers, intelligent pigs, cable sensors etc.

2) *Software-Based Systems*

This refers to models that are driven by software packages to detect and localize a leak. It could be flow/pressure deviation based, pressure point analysis based etc.

3) *Biological/Physical Based Systems*

This refers to models that are driven by physical inspection by humans, dogs etc. to detect and localize a leak.

We go on to further explain some of the leak detection methods.

1) Acoustic Leak Detection

This method is based on the fact that when a leak happens, it produces an acoustic noise around the place of leakage. With the presence of acoustic sensors installed outside the pipe track, these internal noise level can be detected. Normally a baseline of specific features is created and then the self-similarity of this signal is continuously analysed. When a leak occurs, low acoustic signals are detected and investigated. An alarm will be activated if the signal features differ from the baseline.

Advantages

a)  Since sensors are placed outside the pipeline, installation or calibration do not require a shutdown

b)  Interrogation techniques can be used to obtain the leak location.

Limitations

a)  Extra work and costs are incurred.
b)  For high flow rates, the background noise will mask the noise of a leak.

2)  Fibre Optic Sensors

The fibre optic sensing leak detection methods rely on the installation of a fibre optic cable along the pipelines. When a leak occurs in a pipeline, the pipeline commodity touches the fibre cable causing a change in temperature of the cable. This change in temperature is then measured and used to detect and/or locate a leak.

Advantages

a)  It can quickly detect and locate a leak.
b)  It does not require shut down during installation.

Limitations

a)  Extra work and costs are incurred.
b)  It is highly sensitive to environmental conditions.

3)  Balancing Methods

Balancing methods are software based methods that monitor process values and uses them to detect a leak. They could be:

a)  Line Balance Methods

This monitors the difference in actual inflow and outflow and uses a positive balance as a leak detection. Changes in pipeline line pack are ignored and no attempt is made to convert flows to standard conditions.

Advantages

i) It is simple to use and understand.

ii) Extra work and costs are not incurred

Disadvantage

i) It is susceptible to a lot of false alarms.

ii) It cannot determine leak location.

b) Volume Balance Methods

This is like line balance models except that flows are converted to standard conditions.

Advantages

i) It is also simple to use and understand.

ii) Extra work and costs are not incurred

Disadvantages

i) It is susceptible to a lot of false alarms.

ii) It cannot determine leak location.

c) Modified Volume Balance Methods

This is similar to volume balance methods except that an effort is made to compensate for the pipeline line pack.

Advantages

i) It is more accurate.

ii) Little or no extra work/cost is incurred in installation.

Disadvantages

i) The methodology by which the line pack is accounted for is very subjective.

ii) Its leak location accuracy is very low.

d) Compensated Mass Balance Methods

In this methods, modified volume balance is enhanced by keeping track of batches in the pipeline and estimating inventory of each batch based on batch bilk modulus and temperature modulus.

Advantages

i) It is much more accurate than the previous models.

ii) Its methodology is much more objective.

Disadvantage

i) It is complex and may be computationally expensive

e) Real-Time Model Methods

A real-time model (usually referred to as Real-Time Transient Model (RTTM)) is a hydraulic model that stimulates flow in a pipeline using physical laws of conservation (mass, momentum, energy), extensive pipeline parameters (length, diameter, terrain, line equipment etc.) and fluid properties (density, viscosity, bulk modulus etc.).

Since the hydraulic model assumes no leaks, a leak will cause the simulated flow parameter and the expected flow parameter to vary. This deviation is then analysed to detect the presence of a leak, its size, and location. The analysis could take place in so many ways but are usually done as a deviation analysis.

Advantages

i) Like other software-based models, it depends on existing flow data.

ii) It can easily detect leak size and location.

iii) It can provide secondary functions such as batch tracking, pig tracking etc.

Disadvantages

i) Its accuracy is highly dependent on the quality of LDS architecture

ii) It requires a lot of data to be useful

iii) It can be computationally expensive and can become really complex.

f) Statistical Analysis

These methods use advanced statistical techniques such as Bayesian theory, likelihood-Ratio Test, Sequential Probability-Ratio Test etc. to analyse flow rate, pressure and temperature measurements of a pipeline and detecting a leak when there is a statistical variation in these measurements. Recently, it is also used alongside RTTM in a detection method called Extended Real-Time Transient Model (E-RTTM)

Advantages

i)   It greatly reduces the rate of false alarms.

ii)  It is suitable for real-time applications.

iii) It can be used in complex pipe systems.

Disadvantages

i)   It is susceptible to interference of noise.

ii)  It can be complex and also requires a lot of data to work well enough.

### 2.1.3 Leak Detection Performance Metrics.

While the performance of an LDS is on highly divergent factors, the key elements defining it can be defined as follows (as defined by API, 1995b):

a)   Sensitivity

This is defined as the composite measure of the size of a leak that a system is capable of detecting, and the time required for the system to issue an alarm in the event that a leak of that size occurs.

b)   Reliability

This is a measure of the ability of Leak Detection Systems (LDS) to render accurate decisions about the possible existence of a leak on a pipeline

c)   Accuracy

This is defined as a measure of the LDS performance related to estimation parameters such as leak flow rate, total volume lost and leak location.

d)   Robustness

This is defined as a measure of an LDS's ability to continue to function and provide useful information even under changing conditions of pipeline operations

The Alaska Department of Environmental Conservation defined reliability and accuracy to be as follows:

a) Reliability is the probability of detecting a leak, given that a leak does in fact exists, and the probability of incorrectly detecting a leak, given no leak occurred

b) A system is said to be accurate if it can estimate parameters such as leak flow rate, leak location etc. within an acceptable degree of tolerance as defined by the pipeline controller/company.

## 2.2 Intelligent Models

A computer model is said to have computational intelligence or simply, intelligent if it has the ability to learn a specific task from data or experimental observations. They are nature inspired and try to address complex real-world problems in which mathematical or traditional modelling could be useless due to the following reasons:

i) The problem might be too complex for mathematical reasoning

ii) The problem might contain a level of uncertainty

iii) The problem might be stochastic in nature

Computational intelligence has very close ties to soft computing (some believe they are the synonyms). Soft computing is the use of inexact solutions to computational hard task for which no known algorithm can compute an exact solution in polynomial time. It allows for uncertainty, imprecision, partial truths, approximations etc. Its role model is the human mind. The principal components include machine learning, fuzzy logic, evolutionary computation and Bayesian network (Wikipedia).

Computing intelligence (CI) and Soft Computing are differentiated from Artificial Intelligence (AI) and Hard Computing. In 1994, Bezdek referred to CI as a subset of AI. While they have similar long-term goals (reaching general intelligence), how they go about it is different with a significant amount of overlap.

There are quite a number of intelligent models, each optimized for specific applications. For example, convoluted neural networks (CNN) are very good at image identification, recurrent neural networks (RNN) are adept at predicting music notes, decision trees are great at predicting complex decisions etc.

### 2.2.1 Generalized Linear Models

These are methods usually intended for regression which the target value is expected to be a linear combination of the input variables. In mathematical notion:

$$\hat{y}(w,x) = w_0 + w_1 x_1 + \ldots + w_p x_p \qquad 2.1$$

While some argue that linear models cannot be explicitly called intelligent models, it can be seen that by the definition of computational intelligence, they can be said to be intelligent as they all can draw an inference from both experimental and real data.

There are various kinds of linear models but we will be focusing on just some of the most widely used ones for regression problems: Linear Regression (LR), Ridge Regression (RR), Lasso Regression (LAR), and Bayesian Regression (BR).

1) <u>Linear Regression</u>

This is the most basic approach toward a regression problem. While basic, it is a very powerful too in fitting many complex real-life problems and can be tuned into a polynomial regression to account for complex problems. It is simply defined as the linear modelling of the relationship between a dependent variable and one or more independent variable. It is referred to as a simple/ordinary linear regression (OLR) when the independent variable is one or multiple linear regression (MLR) when the independent variable is more than one.

A linear regression model fits a linear model with coefficients $w = (w_1 \ldots w_n)$ to minimize the residual sum of squares between the observed responses in the dataset, and the responses predicted by the model. Mathematically, linear regression solves the problem of the form:

$$\text{Min } \|X_w - y\|^2 \qquad \qquad 2.2$$

2) <u>Ridge Regression</u>

Ridge regression addresses some of the problems of LR by imposing a penalty on the size of the coefficients. These coefficients, usually referred to as ridge coefficients, minimizes a penalized residual sum of squares:

$$\text{Min } \|X_w - y\|^2 + \alpha \|w\|^2 \qquad \qquad 2.3$$

3) <u>Lasso Regression</u>

Lasso regression which refers to least absolute shrinkage and selection operator is a regression analysis methods that performs both variable selection and regularization in order to enhance the prediction accuracy and interpretability of the statistical model it produces (Wikipedia).

While ridge regression shrinks the size of the coefficients to as little as possible, Lasso regression is able to set some of these coefficients to zero by forcing the sum of the absolute value of the regression coefficients to be less than a fixed value.

Mathematically, the objective function is to minimize:

$$\min_{w} \frac{1}{2n_{samples}} ||Xw - y||_2^2 + \alpha ||w||_1$$

2.4

4) <u>Bayesian Regression</u>

Bayesian regression is an approach to linear regression in which the statistical analysis is undertaken within the Bayesian Interface (Figure 2.2). It is used to include regularization parameters in the estimation parameter. This is not set in hand, rather it is tuned to the data at hand

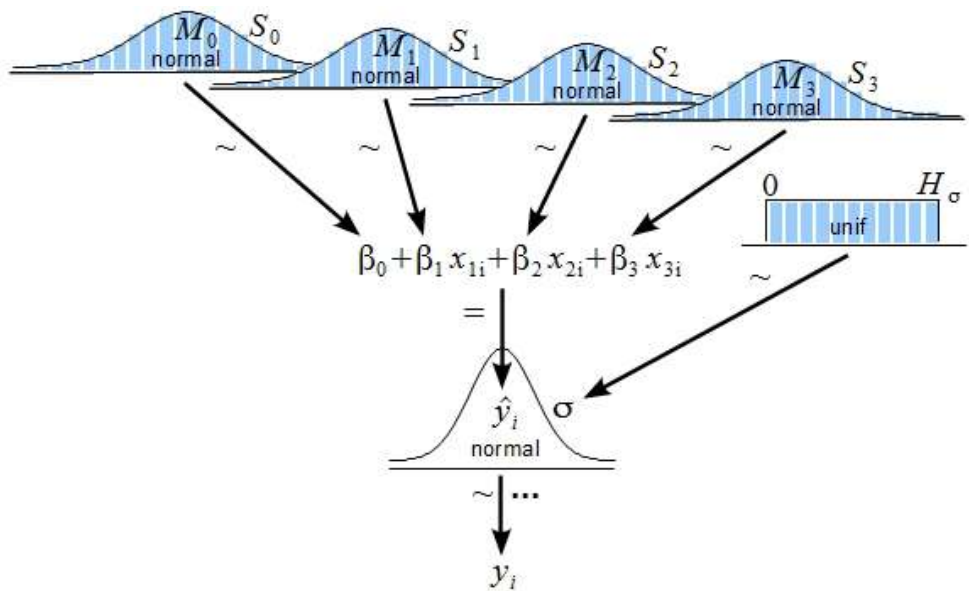

Figure 2.2 Bayesian Regression

Consider a data set of examples of input vectors $\{x_n\}_{n=1}^{N}$ along with corresponding targets $\mathbf{t} = \{t_n\}_{n=1}^{N}$ (this can be easily extended to any number of input vectors). For regression, we generally assume that the targets are some noisy realization of an underlying functional relationship $y(\mathbf{x})$ that we wish to estimate so that

$$t_n = y(\mathbf{x}_n; \mathbf{w}) + \epsilon_n$$

2.5

where $e_n$ is an additive noise process in which the values n are integers, and **w** is a vector of adjustable parameters or 'weights'.

One class of candidate functions for y(**x**, **w**) is given by

$$y(\mathbf{x}; \mathbf{w}) = \sum_{i=1}^{M} w_i \phi_i(\mathbf{x}) = \mathbf{w}^T \boldsymbol{\phi}(\mathbf{x})$$
, 2.6

Which represents a linearly-weighted sum of *M* nonlinear fixed basis functions denoted by φ(**x**) = (φ$_1$(**x**), φ2(**x**)... φ$_M$(**x**))$^T$. Models of these the (2) are known as *linear* models.

Classical (non-Bayesian) techniques use some form of 'estimator' to determine a specific value for the parameter vector **w**. One of the simplest examples is the sum-of-squares error function defined by

$$E(\mathbf{w}) = \frac{1}{2} \sum_{n=1}^{N} |y(\mathbf{x}_n; \mathbf{w}) - t_n|^2$$
2.7

Where the factor of 1/2 is included for later convenience. Minimizing this error function with respect to **w** leads to an estimate **w*** which can be used to make predictions for new values of **x** by evaluating *y*(**x**; **w***).

In a Bayesian approach, the uncertainty in **w** is modelled through a probability distribution *p* (**w**). Observations of data points modify this distribution by virtue of Bayes' theorem, with the effect of the data being mediated through the likelihood function.

Specifically, we define a prior distribution *p* (**w**) which expresses our uncertainty in **w** taking account of all information aside from the data itself, and which, without loss of generality, can be written in the form

$$P(\mathbf{w}|\alpha) \propto \exp\{-\alpha \Omega(\mathbf{w})\}$$ 2.8

Where *α* can again be regarded as a hyperparameter. As a specific example, we might choose a Gaussian distribution for *p* (**w**|*α*) of the form

$$p(\mathbf{w}|\alpha) = \left(\frac{\alpha}{2\pi}\right)^{M/2} \exp\left\{-\frac{\alpha}{2}\|\mathbf{w}\|^2\right\}$$
2.9

We can now use Bayes' theorem to express the posterior distribution for **w** as the product of the prior distribution and the likelihood function

$$P(\mathbf{w}|\mathbf{t}, \alpha, \sigma 2) \propto p(\mathbf{w}|\alpha) L(\mathbf{w})$$ 2.10

Where *L* (**w**) = *p* (**t**|**w**, *σ*2).

In a Bayesian treatment, we make predictions by integrating with respect to the posterior distribution of **w**. If we are given a new value of **x** then the predictive distribution for *t* is obtained from the sum and product rules of probability by marginalizing over **w**

$$p(t|\mathbf{t},\alpha,\beta) = Z\, p(\mathbf{w}|\mathbf{t},\alpha,\sigma^2) p(t|\mathbf{w},\sigma^2) d\mathbf{w}. \qquad 2.11$$

### 2.2.2 Decision Trees

Decision trees can be described also as the combination of mathematical and computational techniques to aid the description, categorization, and generalization of a given set of data. Data comes in records of the form:

$$(\mathbf{x}, Y) = (x_1, x_2, x_3, \ldots, x_k, Y)$$

The dependent variable, Y, is the target variable that we are trying to understand, classify or generalize. The vector x is composed of the features, $x_1$, $x_2$, $x_3$ etc., that are used for that task. Decision tree learning is the construction of a decision tree from class-labelled training tuples.

A decision tree is a flow-chart-like structure, where each internal (non-leaf) node denotes a test on an attribute, each branch represents the outcome of a test, and each leaf (or terminal) node holds a class label (Figure 2.3). The topmost node in a tree is the root node.

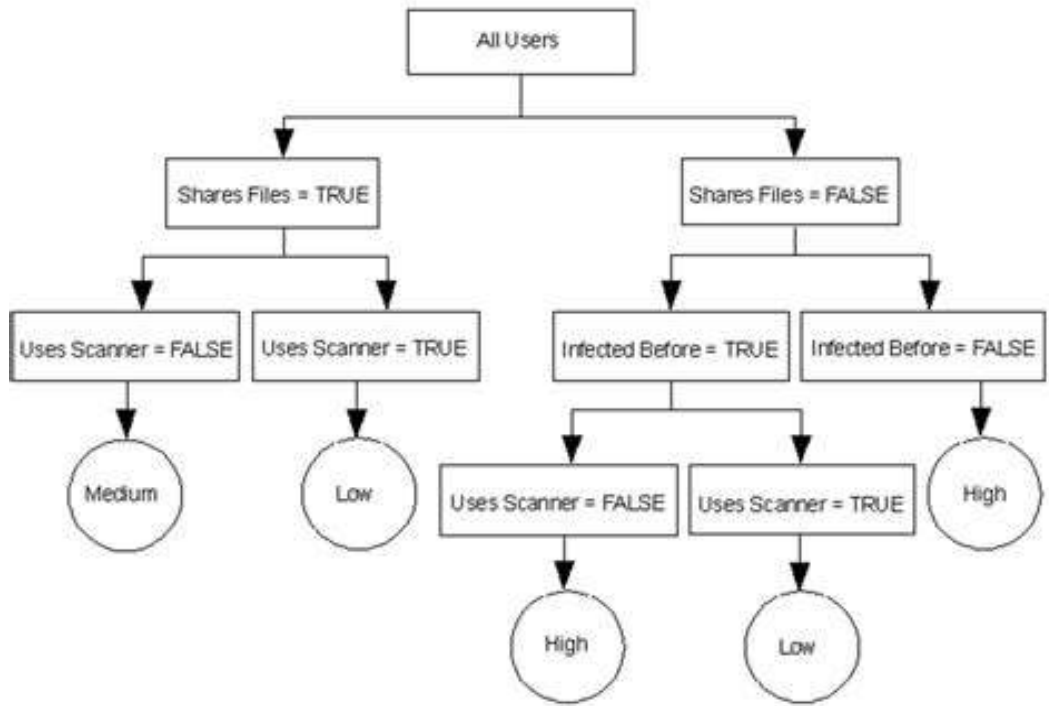

Figure 2.3 A Typical Decision Tree

There are many specific decision-tree algorithms. Notable ones include:

- ID3 (Iterative Dichotomiser 3)
- C4.5 (successor of ID3)
- CART (Classification And Regression Tree)

Algorithms for constructing decision trees usually work top-down, by choosing a variable at each step that best splits the set of items. Different algorithms use different metrics for measuring "best". These generally measure the homogeneity of the target variable within the subsets. Some examples are given below. These metrics are applied to each candidate subset, and the resulting values are combined (e.g., averaged) to provide a measure of the quality of the split.

- Gini impurity

Used by the CART (classification and regression tree) algorithm for classification trees, Gini impurity is a measure of how often a randomly chosen element from the set would be incorrectly labelled if it was randomly labelled according to the distribution of labels in the subset. The Gini impurity can be computed by summing the probability of a mistake in categorizing that item. It reaches its minimum (zero) when all cases in the node fall into a single target category.

To compute Gini impurity for a set of items with J classes, suppose i £ {1,2,…,J}, and let pi be the fraction of items labelled with class i in the set.

$$I_G(p) = \sum_{i=1}^{J} p_i \sum_{k \neq i} p_k = \sum_{i=1}^{J} p_i(1 - p_i) = \sum_{i=1}^{J} (p_i - p_i^2) = \sum_{i=1}^{J} p_i - \sum_{i=1}^{J} p_i^2 = 1 - \sum_{i=1}^{J} p_i^2 \qquad 2.12$$

- Information Gain

Used by the ID3, C4.5 and C5.0 tree-generation algorithms. Information gain is based on the concept of entropy from information theory.

Entropy is defined as below:

$$H(T) = I_E(p_1, p_2, \ldots, p_J) = -\sum_{i=1}^{J} p_i \log_2 p_i \qquad 2.13$$

where $p_1, p_2, \ldots$ are fractions that add up to 1 and represent the percentage of each class present in the child node that results from a split in the tree.

$$\overbrace{IG(T,a)}^{\text{Information Gain}} = \overbrace{H(T)}^{\text{Entropy(parent)}} - \overbrace{H(T|a)}^{\text{Weighted Sum of Entropy(Children)}}$$
$$= -\sum_{i=1}^{J} p_i \log_2 p_i - \sum_{a} p(a) \sum_{i=1}^{J} -Pr(i|a) \log_2 Pr(i|a) \qquad 2.14$$

Information gain is used to decide which feature to split on at each step in building the tree. Simplicity is best, so we want to keep our tree small. To do so, at each step we should choose the split that results in the purest daughter nodes. A commonly used measure of purity is called information which is measured in bits. For each node of the tree, the information value "represents the expected amount of information that would be needed to specify whether a new instance should be classified yes or no, given that the example reached that node".

- Variance reduction

Introduced in CART, variance reduction is often employed in cases where the target variable is continuous (regression tree), meaning that use of many other metrics would first require discretization before being applied.

Advantages

- Simple to understand and interpret.
- Able to handle both numerical and categorical data.
- Requires little data preparation. Other techniques often require data normalization. Since trees can handle qualitative predictors, there is no need to create dummy variables.
- Possible to validate a model using statistical tests.
- Performs well with large datasets.
- Mirrors human decision making more closely than other approaches.

Limitations

- Trees can be very non-robust. A small change in the training data can result in a large change in the tree and consequently the final predictions
- The problem of learning an optimal decision tree is known to be NP-complete under several aspects of optimality and even for simple concepts.
- Decision-tree learners can create over-complex trees that do not generalize well from the training data.

### 2.2.3 K-Nearest Neighbours

The k-nearest neighbour algorithm (k-NN) is a non-parametric method used for classification and regression. In both cases, the input consists of the k closest training examples in the feature space. The output depends on whether k-NN is used for classification or regression. In k-NN classification, the output is a class membership. An object is classified by a majority vote of its neighbours, with the object being assigned to the class most common among its k nearest neighbours (k is a positive integer, typically small). If k = 1, then the object is simply assigned to the class of that single nearest neighbour.

In k-NN regression, the k-NN algorithm is used for estimating continuous variables. One such algorithm uses a weighted average of the k nearest neighbours, weighted by the inverse of their distance. This algorithm works as follows:

- Compute the distance from the query example to the labelled examples using a distance function (Figure 2.4).

*Figure 2.4 Examples of Distance Functions*

- Order the labelled examples by increasing distance.
- Find a heuristically optimal number k of nearest neighbours, based on RMSE. This is done using cross-validation.
- Calculate an inverse distance weighted average with the k-nearest multivariate neighbours

### 2.2.3 Random Forest

The random forest model is a type of additive model that makes predictions by combining decisions from a sequence of base models. More formally we can write this class of models as:

$$G(x) = f_0(x) + f_1(x) + f_2(x) + ... \qquad 2.15$$

Where the final model g is the sum of simple base models fi.

Here, each base classifier is a simple decision tree. This broad technique of using multiple models to obtain better predictive performance is called model ensemble. In random forests, all the base models are constructed independently using a different subsample of the data.

The training algorithm for random forests applies the general technique of bootstrap aggregating, or bagging, to tree learners. Given a training set $X = x_1... x_n$ with responses $Y = y_1... y_n$, bagging repeatedly (B times) selects a random sample with replacement of the training set and fits trees to these samples.

This bootstrapping procedure leads to better model performance because it decreases the variance of the model, without increasing the bias. This means that while the predictions of a single tree are highly sensitive to noise in its training set, the average of many trees is not, as long as the trees are not correlated. Simply training many trees on a single training set would give strongly correlated trees (or even the same tree many times, if the training algorithm is deterministic); bootstrap sampling is a way of de-correlating the trees by showing them different training sets.

Additionally, an estimate of the uncertainty of the prediction can be made as the standard deviation of the predictions from all the individual regression trees on x':

$$\sigma = \sum b = 1 \ B \ (f b \ (x\ ') - f\ \hat{}\ ) \ 2 \ B - 1 \ . \qquad 2.16$$

The number of samples/trees, B, is a free parameter. Typically, a few hundred to several thousand trees are used, depending on the size and nature of the training set. An optimal number of trees B can be found using cross-validation, or by observing the out-of-bag error: the mean prediction error on each training sample $x_i$, using only the trees that did not have $x_i$ in their bootstrap sample. The training and test error tend to level off after some number of trees have been fit.

Random forests differ in only one way from this general scheme: they use a modified tree learning algorithm that selects, at each candidate split in the learning process, a random subset of the features. This process is sometimes called "feature bagging". The reason for doing this is the correlation of the trees in an ordinary bootstrap sample: if one or a few features are very strong predictors for the response variable (target output), these features will be selected in many of the B trees, causing them to become correlated.

### 2.2.4 Boosting Models

Boosting is a machine learning ensemble meta-algorithm for primarily reducing bias, and also variance in supervised learning, and a family of machine learning algorithms that convert weak learners to strong ones.

Boosting algorithms consist of iteratively learning weak classifiers with respect to a distribution and adding them to a final strong classifier. When they are added, they are typically weighted in some way that is usually related to the weak learners' accuracy. After a weak learner is added, the data are reweighted. Thus, future weak learners focus more on the examples that previous weak learners misclassified.

1) AdaBoost

Adaboost was introduced in 1995 by Freund and Schrapine and it is perhaps the most popular boosting model there is. Its core principle is to fit a sequence of weak learners on repeatedly modified versions of the data. The predictions from all of these weak classifiers are then combined to produce the final prediction.

The data modification at each iteration consists of applying weights w1, w2… wn to each of the training samples. Initially, each of these weights is set to 1/N. For each successive training, the learning algorithm does a reweight. Training examples that were incorrectly predicted have an increased weight while the others have a lower one. Thus, each successive weak learner is forced to concentrate on the missed training examples.

2) Gradient Boosting

Gradient boosting involves three elements:

- A loss function to be optimized.
- A weak learner to make predictions.
- An additive model to add weak learners to minimize the loss function.

i) Loss Function

The loss function used depends on the type of problem being solved.

It must be differentiable, but many standard loss functions are supported and you can define your own. For example, regression may use a squared error and classification may use logarithmic loss.

A benefit of the gradient boosting framework is that a new boosting algorithm does not have to be derived for each loss function that may want to be used, instead, it is a generic enough framework that any differentiable loss function can be used.

ii) Weak Learner

Decision trees are used as the weak learner in gradient boosting. Specifically, regression trees are used that output real values for splits and whose output can be added together, allowing subsequent models outputs to be added and "correct" the residuals in the predictions.

iii) <u>Additive Model</u>

Trees are added one at a time, and existing trees in the model are not changed. A gradient descent procedure is used to minimize the loss when adding trees. Traditionally, gradient descent is used to minimize a set of parameters, such as the coefficients in a regression equation or weights in a neural network. After calculating error or loss, the weights are updated to minimize that error.

Instead of parameters, we have weak learner sub-models or more specifically decision trees. After calculating the loss, to perform the gradient descent procedure, we must add a tree to the model that reduces the loss (i.e. follow the gradient). We do this by parameterizing the tree, then modify the parameters of the tree and move in the right direction by reducing the residual loss.

The output for the new tree is then added to the output of the existing sequence of trees in an effort to correct or improve the final output of the model. A fixed number of trees are added or training stops once loss reaches an acceptable level or no longer improves on an external validation dataset.

<u>Improvements to Basic Gradient Boosting</u>

Gradient boosting is a greedy algorithm and can over-fit a training dataset quickly. It can benefit from regularization methods that penalize various parts of the algorithm and generally improve the performance of the algorithm by reducing overfitting.

There are four common enhancements to basic gradient boosting:

i) <u>Tree Constraints</u>

It is important that the weak learners have skill but remain weak. There are a number of ways that the trees can be constrained. A good general heuristic is that the more constrained tree creation is, the more trees you will need in the model, and the reverse, where less constrained individual trees, the fewer trees that will be required.

Below are some constraints that can be imposed on the construction of decision trees:

- The number of trees, generally adding more trees to the model can be very slow to over-fit. The advice is to keep adding trees until no further improvement is observed.
- Tree depth, deeper trees are more complex trees and shorter trees are preferred. Generally, better results are seen with 4-8 levels.

- The number of nodes or number of leaves, like depth, this can constrain the size of the tree, but is not constrained to a symmetrical structure if other constraints are used.
- Number of observations per split imposes a minimum constraint on the amount of training data at a training node before a split can be considered
- Minimum improvement to loss is a constraint on the improvement of any split added to a tree.

ii) Weighted Updates

The predictions of each tree are added together sequentially. The contribution of each tree to this sum can be weighted to slow down the learning by the algorithm. This weighting is called a shrinkage or a learning rate.

The effect is that learning is slowed down, in turn require more trees to be added to the model, in turn taking longer to train, providing a configuration trade-off between the numbers of trees and learning rate. It is common to have small values in the range of 0.1 to 0.3, as well as values less than 0.1.

iii) Stochastic Gradient Boosting

A big insight into bagging ensembles and random forest was allowing trees to be greedily created from subsamples of the training dataset. This same benefit can be used to reduce the correlation between the trees in the sequence in gradient boosting models. This variation of boosting is called stochastic gradient boosting.

At each iteration, a subsample of the training data is drawn at random (without replacement) from the full training dataset. The randomly selected subsample is then used, instead of the full sample, to fit the base learner.

a. Penalized Gradient Boosting

Additional constraints can be imposed on the parameterized trees in addition to their structure. Classical decision trees like CART are not used as weak learners, instead a modified form called a regression tree is used that has numeric values in the leaf nodes (also called terminal nodes). The values in the leaves of the trees can be called weights in some literature.

As such, the leaf weight values of the trees can be regularized using popular regularization functions, such as:

- L1 regularization of weights.
- L2 regularization of weights.
- The additional regularization term helps to smooth the final learnt weights to avoid over-fitting. Intuitively, the regularized objective will tend to select a model employing simple and predictive functions.

### 2.2.5 Support Vector Machines

A Support Vector Machine (SVM) is a discriminative classifier formally defined by a separating hyperplane. In other words, given labelled training data (supervised learning), the algorithm outputs an optimal hyperplane which categorizes new examples.

The numeric input variables (x) in your data (the columns) form an n-dimensional space. For example, if you had two input variables, this would form a two-dimensional space. A hyperplane is a line that splits the input variable space.

In SVM, a hyperplane is selected to best separate the points in the input variable space by their class, either class 0 or class 1. In two-dimensions, you can visualize this as a line and let's assume that all of our input points can be completely separated by this line.

For example:

$$B_0 + (B_1 * X_1) + (B_2 * X_2) = 0 \qquad 2.17$$

Where the coefficients ($B_1$ and $B_2$) that determine the slope of the line and the intercept ($B_0$) are found by the learning algorithm, and $X_1$ and $X_2$ are the two input variables

The SVM algorithm is implemented in practice using a kernel. Although SVMs were originally developed for classification problems; they can also be extended to solve nonlinear regression problems with the introduction of the ε-insensitive loss function.

In support vector regression, the input x is first mapped into a higher-dimensional feature space by the use of a kernel function, and then a linear model is constructed in this feature space. The kernel functions often used in SVM include linear, polynomial, radial basis function, and sigmoid function.

The quality of estimation is measured by a loss function known as ε-insensitive loss function. The advantages of SVM and support vector regression include that they can be used to avoid the difficulties of using linear functions in the high-dimensional feature space, and the optimization problem is transformed into dual convex quadratic programs.

In the case of regression, the loss function is used to penalize errors that are greater than the threshold ε. Such loss functions usually lead to the sparse representation of the decision rule, giving significant algorithmic and representational advantages.

### 2.2.6 Neural Networks

ANNs are processing devices (algorithms or actual hardware) that are loosely modelled after the neuronal structure of the mammalian cerebral cortex but on much smaller scales. A large ANN might have hundreds or thousands of processor units, whereas a mammalian brain has billions of neurons with a corresponding increase in the magnitude of their overall interaction and emergent behaviour.

Neural networks are typically organized in layers. Layers are made up of a number of interconnected 'nodes' which contain an 'activation function'. Patterns are presented to the network via the 'input layer', which communicates to one or more 'hidden layers' where the actual processing is done via a system of weighted 'connections'. The hidden layers then link to an 'output layer' where the answer is output as shown in Figure 2.5.

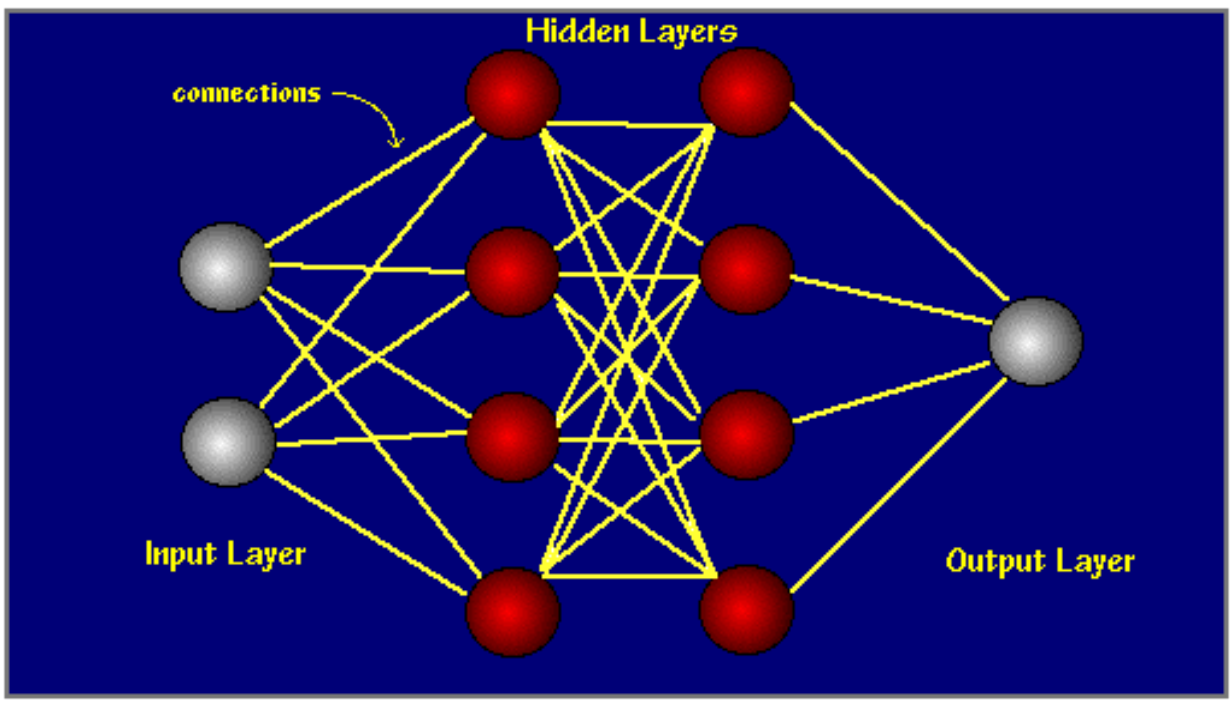

Figure 2.5 an ANN Architecture

The type of model determines how the network connects the predictors to the targets through the hidden layer(s). The multilayer perceptron (MLP) allows for more complex relationships at the possible cost of increasing the training and scoring time. The radial basis function (RBF) may have lower training and scoring times, at the possible cost of reduced predictive power compared to the MLP.

Information flows through a neural network in two ways. When it's learning (being trained) or operating normally (after being trained), patterns of information are fed into the network via the input units, which trigger the layers of hidden units, and these in turn arrive at the output units. This common design is called a feedforward network.

Not all units "fire" all the time. Each unit receives inputs from the units to its left, and the inputs are multiplied by the weights of the connections they travel along. Every unit adds up all the inputs it receives in this way and (in the simplest type of network) if the sum is more than a certain threshold value, the unit "fires" and triggers the units it's connected to (those on its right).

For a neural network to learn, there has to be an element of feedback involved. Neural networks learn things through a feedback process called backpropagation (sometimes abbreviated as "backprop"). This involves comparing the output a network produces with the output it was meant to produce, and using the difference between them to modify the weights of the connections between the units in the network, working from the output units through the hidden units to the input units—going backward, in other words. In time, backpropagation causes the network to learn, reducing the difference between actual and intended output to the point where the two exactly coincide, so the network figures things out exactly as it should.

Types of ANN includes but not limited to:

i) <u>A Perceptron Network</u>

This neural network is one of the simplest forms of ANN, where the data or the input travels in one direction. The data passes through the input nodes and exit on the output nodes. This neural network may or may not have the hidden layers. In simple words, it has a front propagated wave and no backpropagation by using a classifying activation function usually.

ii) <u>Radial basis function Neural Network</u>

Radial basis functions consider the distance of a point with respect to the center. RBF functions have two layers, first where the features are combined with the Radial Basis Function in the inner layer and then the output of these features are taken into consideration while computing the same output in the next time-step which is basically a memory.

iii) <u>Kohonen Self Organizing Neural Network</u>

The objective of a Kohonen map is to input vectors of arbitrary dimension to discrete map comprised of neurons. The map needs to be trained to create its own organization of the training data. It comprises of either one or two dimensions. When training the map the location of the neuron remains constant but the weights differ depending on the value. This self-organization process has different parts, in the first phase every neuron value is initialized with a small weight and the input vector. In the second phase, the neuron closest to the point is the 'winning neuron' and the neurons connected to the winning neuron will also move towards the point. The distance between the point and the neurons is calculated by the Euclidean distance, the neuron with the least distance wins. Through the iterations, all the points are clustered and each neuron represents each kind of cluster.

iv) <u>Recurrent Neural Network (RNN)</u>

The Recurrent Neural Network works on the principle of saving the output of a layer and feeding this back to the input to help in predicting the outcome of the layer. Here, the first layer is formed similarly to the feed forward neural network with the product of the sum of the weights and the features.

The recurrent neural network process starts once this is computed, this means that from a one-time step to the next each neuron will remember some information it had in the previous time-step. This makes each neuron act like a memory cell in performing computations. In this process, we need to let the neural network to work on the front propagation and remember what information it needs for later use.

Here, if the prediction is wrong we use the learning rate or error correction to make small changes so that it will gradually work towards making the right prediction during the backpropagation.

2.2.6 Neuro-Fuzzy Systems

Fuzzy logic systems propose a mathematical calculus to translate the subjective human knowledge of the real processes. This is a way to manipulate practical knowledge with some level of uncertainty. The fuzzy sets theory was initiated by Lofti Zadeh in 1965. The point of fuzzy logic is to map an input space to an output space, and the primary mechanism for doing this is a list of if-then statements called rules. All rules are evaluated in parallel, and the order of the rules is unimportant.

The fuzzy inference mechanism consists of three stages: in the first stage, the values of the numerical inputs are mapped by a function according to a degree of compatibility of the respective fuzzy sets, this operation can be called fuzzification. In the second stage, the fuzzy system processes the rules in accordance with the firing strengths of the inputs. In the third stage, the resultant fuzzy values are transformed again into numerical values, this operation can be called defuzzification (Figure 2.6).

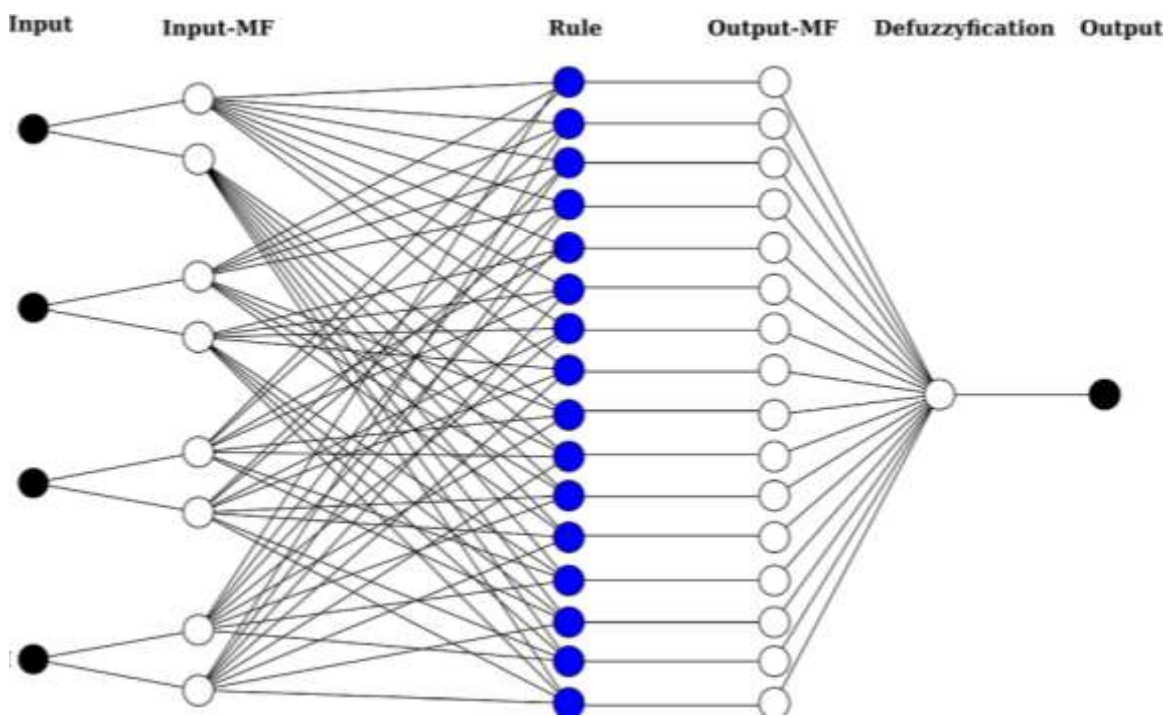

Figure 2.6 A NeuroFuzzy System

Advantages of Fuzzy Logic includes:

- Fuzzy logic is conceptually easy to understand.
- The mathematical concepts behind fuzzy reasoning are very simple. Fuzzy logic is a more intuitive approach without the far-reaching complexity.

- Fuzzy logic is flexible.
- Fuzzy logic is tolerant of imprecise data.
- Fuzzy logic can model nonlinear functions of arbitrary complexity.
- Fuzzy logic can be built on top of the experience of experts.

And its disadvantages are:

- It is incapable to generalize, or either, it only answers to what is written in its rule base.
- It is not robust in relation to the topological changes of the system, such changes would demand alterations in the rule base.
- It depends on the existence of an expert to determine the inference logical rules.
- Neuro-fuzzy systems are the fusion of neural network and fuzzy logic. The neural network helps the fuzzy logic system to generalize and become robust. Similarly, with Neuro-Fuzzy systems, there will be no need for an expert to determine the inference logical rules.

The classes of neuro-fuzzy logic systems include:

- <u>Cooperative Neuro-Fuzzy System</u>

In the cooperative systems, there is a pre-processing phase where the neural networks mechanisms of learning determine some sub-blocks of the fuzzy system. For instance, the fuzzy sets and/or fuzzy rules (fuzzy associative memories or the use of clustering algorithms to determine the rules and fuzzy sets position). After the fuzzy sub-blocks are calculated the neural network learning methods are taken away, executing only the fuzzy system.

- <u>Concurrent Neuro-Fuzzy System</u>

In the concurrent systems, the neural network and the fuzzy system work continuously together. In general, the neural networks pre-processes the inputs (or post-processes the outputs) of the fuzzy system.

- <u>Hybrid Neuro-Fuzzy System</u>

In this category, a neural network is used to learn some parameters of the fuzzy system (parameters of the fuzzy sets, fuzzy rules and weights of the rules) of a fuzzy system in an iterative way. The majority of researchers uses the neuro-fuzzy term to refer only hybrid neuro-fuzzy system.

## 2.3 Literature Review of Leak Detection Systems and Intelligent Models

Belsito et al. (1998) used an artificial neural network to build leak detection systems. This system can detect and locate leaks down to 1% of flow rates in pipelines carrying hazardous materials in about 100s. The package also correctly predicts the leaking segment of a pipeline with a probability of success greater than 50%.

Scott et al. (1999) modelled leaks in multiphase flow lines in deep water. Their model could also be simplified to single-phase gas pipelines.

Carpenter et al. (2003) used a Bayesian Belief Network (BNN) for pipeline leak detection. These disparate sources of information as well as prior probabilities of events as part of an effort to detect leaks, reduce false positive and optimize leak detection threshold.

Caputo (2003) proposed a state estimation technique which uses Artificial Neural Networks (ANN) to monitor the status of piping networks carrying hazardous fluids in order to identify and locate spills and leakages. It was seen that the theoretical capacity of ANN to monitor piping state network was promising.

Ahmad et al. (2003) used a neural network based detection scheme integrating a neural Elman network dynamic predictor – an RNN and a feedforward neural network fault classifier to detect a leak in a palm oil fractionation process. The model was seen to be able to detect a leak of as low as 0.1%.

Da Silva et al. (2005) used a combination of clustering and classification tools for leakage detection in a pipeline. A fuzzy system is used to classify the running mode and identify the operational and process transients. It was applied to a small scale LPG pipeline and the results were very encouraging.

Wang et al. (2006) used transient modelling to detect leaks in both gas and liquid pipelines. The modelling was done as an online observer thus it could auto-study and self-adjust. The model was used on both gas and liquid pipelines and was found to be more accurate for liquid pipelines.

Ferraz et al. (2008) used an ensemble of ANNs to detect a leak in pipelines. The LDS composed of four specialists. The first specialist classified read values into a leak or no leak. The second classified the same input values into different flow conditions. The third detects the leak flow rate while the last detected the leak location.

Casillas et al. (2012) used optimal linear regression models to detect a leak in water distribution networks. The models were then compared using the predicted and actual data obtained from pressure measurements along a time horizon. The results were encouraging in different scenarios of noise and demand pattern.

Mandal et al. (2012) used a leak detection scheme based on rough set theory and support vector machine to overcome the problem of false leak detection in oil pipelines. For the SVM, swarm intelligence technique (Ant Bee Colony) was used for training. The model gave a 95.2% leak detection accuracy.

Afebu et al. (2015) used an integrated system of OLGA simulator and Artificial Neural Network (ANN) to detect a leak in gas pipelines. OLGA simulator was used to simulate no leak and different leak conditions. The ANN was used to identify the simulated leaks. It was seen that the system was for accurate for leak location than leak size.

Gieger et al. (2015) used a combined leak detection of RTTM, Volume balance and Rarefaction to detect a leak and predict their location using pattern recognition techniques. It is was tested on a liquid multi-product pipeline in Germany. The system finds leaks in less than a minute.

Desmet et al (2017) used unsupervised anomaly detection techniques to detect leaks in compressed air systems. A wavelet transform, random forest, clustering, and neural network autoencoder were used. The clustering algorithm was seen to outperform the neural network autoencoder.

Chen et al (2018) used a decision tree and multi-support vector machine to detect leak detection in pipelines. The model was trained with a fault feature vector. The results show that this method competes well in the case of small samples and can greatly improve the neural network method in terms of recognition performance and can be effectively applied to leakage detection in pipelines.

# CHAPTER THREE

# MODEL DEVELOPMENT

## 3.1 Introduction

This project aims to apply the observer design technique to detect leaks in gas pipelines. The problem was treated as a regresso-classification hierarchical problem where the intelligent model acts as a regressor and a modified logistic regression acts as a classifier.

## 3.2 Model Formulation

The Real Time Transient Model is a fluid hydraulics model that replicates the actual pipeline. It is driven on the real time operational data and boundary conditions and is based on the physical laws of conservation of mass, momentum and energy. This method relies on the fact that a leak generates its own transients or signals and the pressure and flow waves produced by the leak propagates to the ends of the pipeline where it generates an imprint on the measured data. (Balda Rivas, et al., 2013)

Typically, a RTTM will calculate fluid flow, pressure and temperature for the entire pipeline based on some field data from the SCADA system: flow, pressure, temperature, density at certain receipt and delivery locations, referred to as boundary conditions.

To develop a model certain assumptions must be made, in this work the following assumptions will be used to model natural gas flow through a pipeline.

1. Pipeline cross sectional area is constant
2. Flow is horizontal
3. The fluid is single phase, homogenous and compressible
4. No chemical reactions between the pipe and the fluid exist

The three governing equation as follows:

Mass: $\quad \dfrac{\partial \rho}{\partial t} + \dfrac{\partial \rho u}{\partial x} = 0 \quad$ 3.1

Momentum: $\quad \dfrac{\partial}{\partial t}(\rho u dx) + \dfrac{\partial}{\partial x}(p + \rho u^2) dx = F_g + F_f \quad$ 3.2

Energy: $\quad C_v \dfrac{\partial}{\partial t} T + C_p u \dfrac{\partial}{\partial x} T = q_h - gu\sin\theta \quad$ 3.3

Adding equations 3.1, 3.2, 3.3:

$$\frac{\partial}{\partial t}(p + \rho_u + T\rho) + \frac{\partial}{\partial x}(\rho u + p + \rho u^2 + T\rho u) = \beta + \gamma \qquad 3.4$$

Where:

$$\beta = -\left(\rho g \sin\theta + \frac{\rho u^{2f}}{2D}\right), \gamma = \frac{1}{c^1}(\rho q_h - \rho g \sin\theta)$$

$$q_h = \frac{4U_h(Ta - T)}{D}$$

Writing equation 3.33 in vector form

$$\frac{\partial \vec{v}}{\partial t} + \frac{\partial \vec{f}}{\partial x} + \vec{D} = 0 \qquad 3.5$$

Where

$$\vec{v} = \begin{pmatrix} \rho \\ \rho u \\ T\rho \end{pmatrix} \vec{F} = \begin{pmatrix} \rho u \\ P + \rho u \\ T\rho u \end{pmatrix} \vec{D} = \begin{pmatrix} 0 \\ \beta \\ \gamma \end{pmatrix}$$

Usually, three forms of techniques are used to solve the equation 3.5. They are:

i)        Explicit Methods
ii)      Implicit Methods
iii)     Methods of characteristics

These techniques are numerical in nature and are said to be model-driven i.e. they are solved as approximations of the equation 3.5 as an extension of analytical solutions of equations using sufficient initial and boundary conditions alongside some other simplifying assumptions. In contrast, intelligent models are said to be data-driven i.e they are solved as approximations of the equation 3.5 as an extension of statistical learning theory using sufficient data that completely describes the equation.

In Chapter two of this project, over 10 intelligent models were discussed. For the purpose of these project, only five of those models will be used. The choice of intelligent models to use in this research was governed by two main criteria:

a)    Their inherent theoretical strength,
b)    Their popularity in recent literature for leak detection or similar problems.

With these criteria in place, the following models were chosen:

i) Decision Tree (DT)

ii) Random forest (RF)

iii) Gradient Boosting (GB)

iv) Support Vector Machine (SVM)

v) Artificial Neural Network (ANN)

### 3.2.1 Decision Trees, Random Forest and Gradient Boosting

In decision trees, we aim to approximate the equation 3.5 such that information variance at any particular node given as equation 3.6 is minimum.

$$I_V(N) = \frac{1}{|S|^2}\sum_{i \in S}\sum_{j \in S}\frac{1}{2}(x_i - x_j)^2 - \left(\frac{1}{|S_t|^2}\sum_{i \in S_t}\sum_{j \in S_t}\frac{1}{2}(x_i - x_j)^2 + \frac{1}{|S_f|^2}\sum_{i \in S_f}\sum_{j \in S_f}\frac{1}{2}(x_i - x_j)^2\right) \quad 3.6$$

where:

Iv(N) is the information variance at node N

S is the number of node

Xi is the value of the parent node

Xj is the value of the child node

Random Forest is simply an ensemble of decision trees where each decision tree is shown a bootstrapped sample of the training population and the aim is to minimize the cost function given by equation 3.6. Whereas a Gradient Boosting Tree is an ensemble of decision trees where each tree is built successively not concurrently and the each successive tree is built to minimize the errors of the previous tree.

### 3.2.2 Support Vector Machine

For a support vector machine, the equation 3.5 can be easily represented as:

$$f(x, w) = Gv_0 + b = 0 \quad 3.7$$

where   G is a kernel matrix

V0 is a matrix of the weighting vectors

b is the bias term

w is the weight parameters.

x is the parameters.

We aim to minimize the error by measuring the empirical error term Remp given by:

$$R^\varepsilon emp(w,b) = (1/L) \sum |y_i - w^T x_i - b|_\varepsilon \qquad 3.8$$

where

$Y_i$ is the target variables

$\varepsilon$ is the Vapnik's $\varepsilon$-insensitivity loss function given as:

$$e(x,y,f) = \max(0, |y - f(x, w)| - \varepsilon).$$

Solving the equation 3.8 by forming primal variables Lagrangian, we arrive at the optimal weight vector of the regression hyperplane given as:

$$w_0 = \sum_{i=1}^{l} (\alpha_i - \alpha_i^*) x_i. \qquad 3.9$$

Where:

$\alpha_i, \alpha_i^*$ are referred to as the Lagragian multipliers.

### 3.2.3 Perceptron Neural Network

For a perceptron neural netwoek, the equation 3.5 can be easily represented as:

$$f(x, w) = \phi(x \cdot w) = \phi\left(\sum_{i=1}^{p} (x_i \cdot w_i)\right) \qquad 3.10$$

Where:

$\Phi$ is an activation function

w is a matrix of weights

x is a matrix of inputs

The aim is to obtain the weights of equation 3.10 such that the mean absolute error is very close to zero. To do this several solvers can be used such as Adam, LBFGS etc. To solve, we perform this, we go through the following:

Forward Pass:

$$X_i = \Phi(W_i x_{i-1}) \tag{3.10a}$$

$$E = \|x_L - t\|^2_2 \tag{3.10b}$$

Backward Pass:

$$\delta_L = (x_L - t) \circ f'_L(W_L x_L - 1) \tag{3.10c}$$

$$\delta_i = W^T_{i+1} \delta_{i+1} \circ f'_i(W_i x_{i-1}) \tag{3.10d}$$

Weight Update:

$$\partial E / \partial W_i = \delta_i x^T_{i-1} \tag{3.10e}$$

$$W_i = W_i - \alpha W_i \circ \partial E / \partial W_i \tag{3.10f}$$

Where E is the error cost function

    $\alpha$ is the learning rate

    W is the matrix of weights

    $\delta$ is the error for each term

It is an iterative procedure and the optimization is usually done using one of the aforementioned solvers.

### 3.3 Leak Detection

This is done by residual analysis. The intelligent model acts as an observer for the pipeline. It predicts the outlet and inlet flowrate using basic operational parameters only. The residuals or leak signature is then analysed using a modified logistic regression model which works in tandem with the intelligent model to detect a leak when a leak index is greater than a particular threshold. To avoid false alarms, the leak alarm is set to be persistent for six minutes before an alarm is issued.

### 3.3.1 Leak Index

The flowrate residuals are monitored for the previous one hour and the number of residuals (a) greater than a particular threshold is counted and used to derive a leak index. The leak index formula is in an exponential form. It is basically a modified form of the sigmoid function. It is given as:

$$f(a) = \frac{1}{1+exp^{-a}} \quad \text{(Sigmoid function)} \quad 3.11$$

$$f(a) = \exp\left(\frac{-1}{1+a^2}\right) \quad \text{(Modified function)} \quad 3.12$$

The modified function was chosen as the leak index function because it makes the model to be more robust to false alarms.

### 3.3.2 Choice of the Threshold

Five threshold choices were considered:

i) A constant number, c.
ii) A constant residual percentage, c% of output flowrate.
iii) The mean absolute error (MAE) of the model.
iv) The root mean squared error (RMSE) of the model.
v) Adjusted forms of the MAE and/or RMSE of the model.

It was seen that for most models the adjusted form of MAE was the most robust to false alarms. Thus the choice of the threshold is given by:

$$threshold = \text{MAE} + 0.01 \quad 3.13$$

### 3.3.3 Leak Localization

To estimate leak location, two approaches can be used:

1) Detection of a rarefaction wave generated by the leak,
2) Analysing the pattern of inconsistencies in the RTTM results.

If the model accurately approximates the pipeline operations and when the leak flow rate is small compared to the actual flow rate, one can obtain an estimate of the leak location using a linearized estimate given as:

$$\text{Fractional leak location} =$$

Flow deviation at the outlet/ (Flow deviation at the outlet + Flow deviation at the inlet)     3.14

(Morgan et al., 2015)

### 3.4 Data Analysis

#### 3.4.1 Introduction

The data acquired will undergo the normal data science pipeline as shown in the figure below.

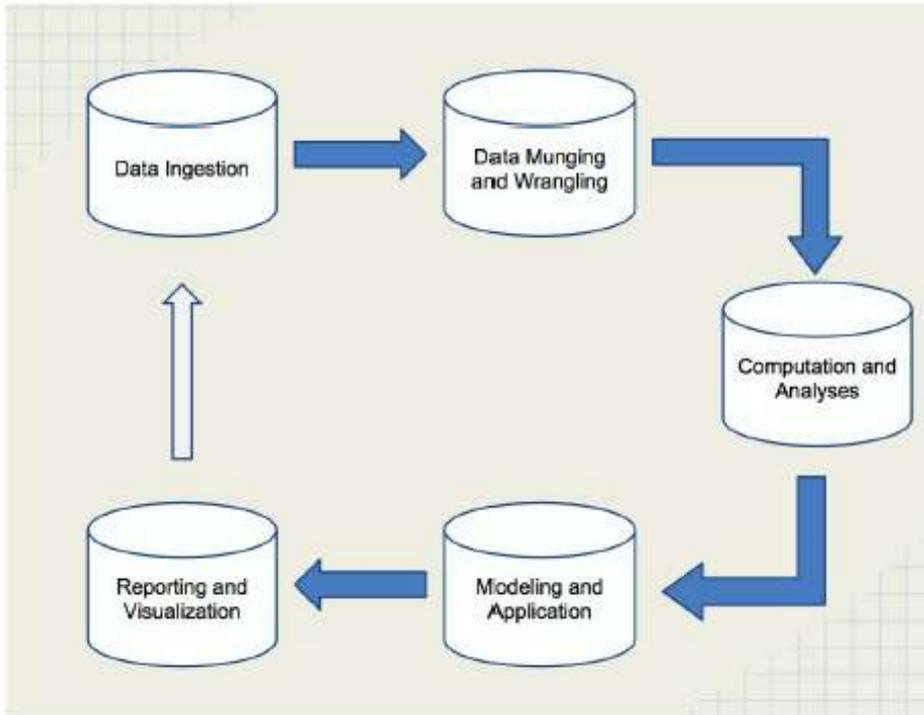

Figure 3.1 Data Analysis Pipeline

#### 3.4.2 Data Acquisition and Ingestion

Pipeline X is a pipeline in the Niger Delta area of Nigeria with a length of about 11km and roughness of 0.002 inches. It has a diameter of 21 inches. Ambient temperature ranges for 20°C - 32°C. The historical data was obtained from SCADA measurements of pressure, temperature and flow rate for both inlet and outlet that came in every two minutes for a period of four weeks resulting in a total of about 21000 data points.

Data ingestion is firstly done using Excel. The field data is then validated with a commercial transient simulator using the pressure measurements as boundary conditions. The transient simulator combines the following equations for the conservation of mass, momentum, and energy alongside some correlations (the particular ones are not made public) such as friction factor correlation, viscosity correlation, gas compressibility correlations etc. to simulate transient flow along a pipeline. The validation was done to see how accurate the data followed theoretical background.

### 3.4.3 Data Munging and Wrangling

This involves making the data useful for computation in the intelligent models. This is done using a Microsoft Excel tool called Power Query. All data points that contained null values were removed from the data set. However, outliers were kept because they could correlate with periods of transients in the gas pipelines.

### 3.4.4 Computation and Analysis

*1) Choice of programming language*

Three programming languages were considered:

i)   JAVA

ii)  Python

iii) MATLAB

Python is the language of choice because of the following reasons:

a) It is an open-source language meaning the results of this research can easily be reproduced or adapted for other similar projects.

b) It has an active community of developers which means previous libraries get improved really quickly for computational efficiency.

c) It is computationally fast.

*2) Choice of Libraries*

The following libraries were used:

a) Numpy and Pandas for storing and cleaning data,

b) Scikit-learn for implementation of intelligent models,

c) Matplotlib and Seaborn for data visualization

### 3.4.5 Modelling and Application

*1) Model Metrics*

The models are trained and tuned using:

a) <u>Root Mean Squared Error</u>

This is a quadratic scoring rule that measures the average magnitude of the error. It's the square root of the average of squared differences between prediction and actual observation. As it is an error score, lower values (as close to zero as possible) is preferred.

$$RMSE = \sqrt{\frac{1}{n} \sum_{j=1}^{n} (y_j - \hat{y}_j)^2}$$

3.15

b) <u>Mean Average Error</u>

MAE measures the average magnitude of the errors in a set of predictions, without considering their direction. It's the average over the test sample of the absolute differences between prediction and actual observation where all individual differences have equal weight. Since it is also an error score, lower values (as close to zero as possible) is preferred.

$$MAE = \frac{1}{n} \sum_{j=1}^{n} |y_j - \hat{y}_j|$$

3.16

c) <u>Coefficient of Determination ($R^2$ Score)</u>

It is a measure that assesses the ability of a model to predict or explain an outcome in the linear regression setting. More specifically, $R^2$ indicates the proportion of the variance in the dependent variable (*Y*) that is predicted or explained by a regressor and the predictor variable (*X*, also known as the independent variable). In general, a high $R^2$ value indicates that the model is a good fit for the data, although interpretations of fit depend on the context of analysis.

$$R^2 = 1 - \frac{\sum (Y_i - \hat{Y})^2}{\sum (Y_i - \bar{Y})^2} \qquad \text{where } \hat{Y} \text{ is predicted value, } \bar{Y} \text{ is mean}$$

3.17

*2) Model Training Method*

Training will be done in a batch mode. However, even after the original training has occurred, additional data obtained from SCADA that has been validated to be a no-leak situation can be used to improve the model to ensure robustness. Batch training was chosen due to hardware constraints. Without these constraints, online training will be preferred.

70% of the data will be used for training while 30% will be used as the test set. Usually, an extra set called the cross-validation set is kept to ensure the model generalizes well.

However, in the model, K-fold Cross-validation is used. In K-fold Cross-validation, depending on the number of folds, a specified percentage of the training set is randomly used to cross-validate k amount of times and the metrics are printed. This is to make the models more robust and avoid sampling error.

3) *Parameter/ Feature Selection*

Usually, feature selection can be done automatically using:

i) Univariate statistics,
ii) Model-based feature selection,
iii) Iterative feature selection,
iv) Expert domain knowledge

In this research, expert domain knowledge and iterative feature selection are used. By literature review (i.e. expert domain knowledge), the basic operational parameters are inlet and outlet pressures, inlet and outlet temperatures, gas composition & inlet and outlet mass/volume flowrates. Other important parameters like z-factor, specific gravity and gas viscosity can be obtained using correlations and these basic operational parameters.

Gas viscosity is estimated using Lee, Gonzalez and Eakin (1966) correlation given below:

$$\mu = K exp(X(\frac{\rho}{62.4})^y) \qquad 3.18$$

Where:

$$K = T^{1.5} \frac{(9.4+0.02M_g)}{10^4(209+19M_g+T)} \qquad 3.19$$

$$X = 3.5 + 0.01M_g - \frac{986}{T} \qquad 3.20$$

$$y = 2.4 - 0.2X \qquad 3.21$$

Gas Compressibility can be obtained by the correlation by Beggs and Brill (Golan and Whitson, 1986) for the calculation of z. It is given below:

$$z = A + (1 - A)e^{-B} + CP_{pr}^{D} \qquad 3.22$$

Where

$$A = 1.39(T_{pr} - 0.92)^{0.5} - 0.36T_{pr} - 0.101$$

$$B = (0.62 - 0.23T_{pr})P_{pr} + \left[\left(\frac{0.066}{T_{pr}-0.86}\right) - 0.037\right]P_{pr}^{2} + \left[\frac{0.32}{10^{9}(T_{pr}-1)}\right]P_{pr}^{6}$$

$$C = (0.132 - 0.32 \log T_{pr})$$

$$D = 10^{K}$$

$$K = 0.3106 - 0.9T_{pr} + 0.1824T_{pr}^{2} \qquad 3.23a\text{-}3.23e$$

Thus, by domain knowledge we have the following features:

- Inlet pressure
- Outlet pressure
- Inlet Flowrate
- Outlet Flowrate
- Inlet Temperature
- Outlet Temperature
- Gas compressibility
- Gas viscosity
- Gas specific gravity

By iterative feature selection (i.e. we plugged in all the features into the model, check for its accuracy and start removing one feature after the other to see which one has the least effect on the model's accuracy in order to make the model more robust), we are left with five features:

- Inlet pressure
- Outlet pressure
- Inlet temperature
- Outlet temperature

- Flowrate (predicted class).

From a petroleum engineering point of view using the general pipeline flow equation, these final features make sense. These features were reprocessed into polynomials forms for GB, RF, SVM, and DT while they were left in their original states for ANN.

### 3.4.6 Model Architecture

Generally, parameters to be tuned for tuning in each model architecture was chosen based on literature and industrial application knowledge on which parameters in a model is most likely to have the greatest effect on the model's strength. The number of parameters to be tuned is limited to 2 for computational efficiency. The range of the tuning was set domain knowledge and 'trial and error'.

1) <u>Gradient Boosting</u>

Grid searching in python was used to pick the best model. The interest tuned parameters were:

- Learning rate (range between .1 and 10)

- Number of estimators (range between 50 and 500)

The criterion for training was the Friedman's mean squared error which is just an improvement on the ordinary mean squared error. The loss function to be optimized was the least squared regression. All other parameters were set to Sklearn's default values.

2) <u>Random Forest</u>

Grid searching in python was used to find the best model. The parameters of interest are the number of estimators (range between 10,100) and the maximum features (one of square root, auto or log choices). The criterion for training was the mean squared error. All other parameters were set to Sklearn's default.

3) <u>Decision Tree</u>

The criterion for training was the Friedman's mean squared error. The parameters of interest are:

- Maximum features (None, log2 or auto)
- Min samples split (2,5 or 10)

All other parameters were set to Sklearn's default.

4) <u>Support Vector Machines</u>

Because of the internal structure of support vector machines, they are highly susceptible to feature scaling. Thus feature scaling was first performed on the train and test set for good accuracy scores. Grid searching was then used to pick the best model. The parameters of interest are the regularization parameter, C (range between .1, 10000) and the kernel. The kernel is set to radial basis function (RBF) or linear kernel. The gamma parameter, another type of regularization parameter, was set to scale while the rest of the parameters were set to Sklearn's default.

5) <u>Artificial Neural Network</u>

The ANN architecture used was Multi-layer Perceptron. Like SVM, the parameters were scaled for ANN. Grid searching was then used to pick the best parameters. The parameters of interest were:

- Number of hidden layers (range of 1 or 2)
- Number of hidden layers per hidden layer (1,20)
- Alpha (between 0.0001 and 10)

The following parameters were set using domain knowledge:

- Maximum iteration=1000
- Momentum = 0.1
- Solver = LBFGS
- Activation function = tanh

All other values were set to Sklearn's default.

### 3.4.7 Reporting and Visualization

The following tools will be used for reporting and visualization:

- Python's Matplotlib library,
- Excel Charts,
- Excel Tables
- Excel Pivot Tables and Pivot Charts.

## 3.5 Leak Detection Performance Metrics

The leak detection performance metrics were chosen based on applicable metrics as defined by API, 1995b. These metrics are:

- Sensitivity (minimum detectable leak and the time required for the system to issue an alarm if such a leak occurs).
- Reliability (the ability of the system to render accurate results as regards the possible existence of a leak on a pipeline)
- Accuracy (a measure of the system to accurately estimate leak flow rate)
- Robustness (a measure of the system to continue to function and provide useful information even under changing conditions of pipeline operations).

Other important industry metrics are

- The average time to leak detection is defined as the time taken for the model to detect a leak of certain leak size regardless of its position.
- Leak localization error is defined the average error in percentage by which the leak detection system overestimates or underestimates the leak position.

# CHAPTER FOUR

## RESULTS AND DISCUSSION

### 4.1 RESULTS

The results for the following are shown in this chapter:

i) Data Ingestion
ii) Data Validation
iii) Model Validation

### 4.1.1 Data Ingestion

Data ingestion is done to understand the kind of boundary conditions the pipeline had. This is important in order to validate the field data against a transient simulator. Figure 4.1 shows the graph of inlet and outlet pressures against time. The plotted points are daily averages. It can be seen that at the inlet, the pressure is always almost constant. This is consistent with the expected results. Figure 4.2 shows the graph of outlet flowrate against time. While, for a certain stretch of time, the outlet flow rate is almost always constant, the varying flow rate shows that the pipeline system is almost always in a mild transient state. There are major transients at some certain times. Figure 4.3 shows the graph of inlet and outlet temperatures. The variation between inlet and outlet temperatures shows that the pipeline is not isothermal and there are significant thermal deviations at some certain times.

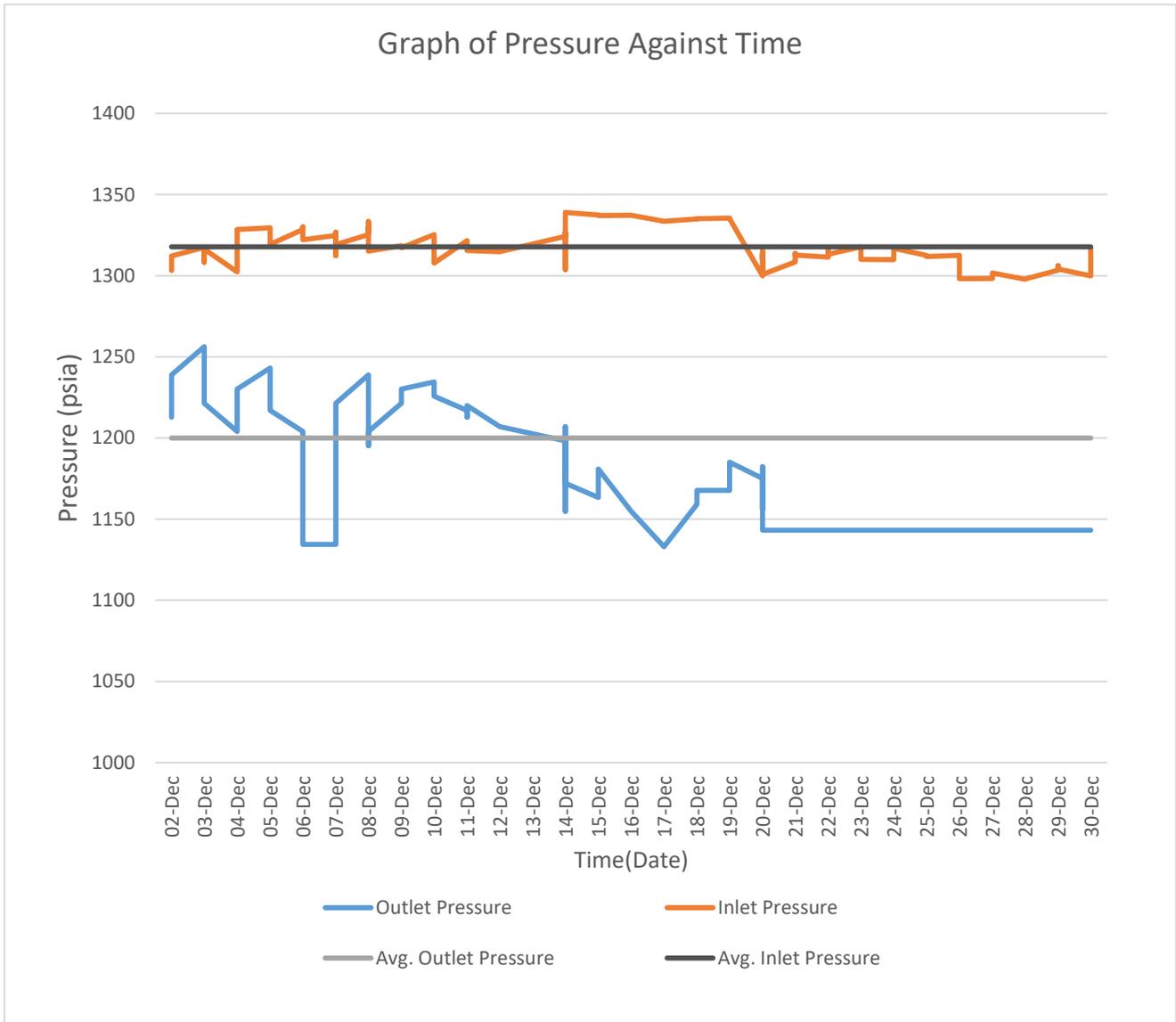

Figure 4.1 Graph of Daily Average Inlet and Outlet Pressure against Time

49 | P a g e

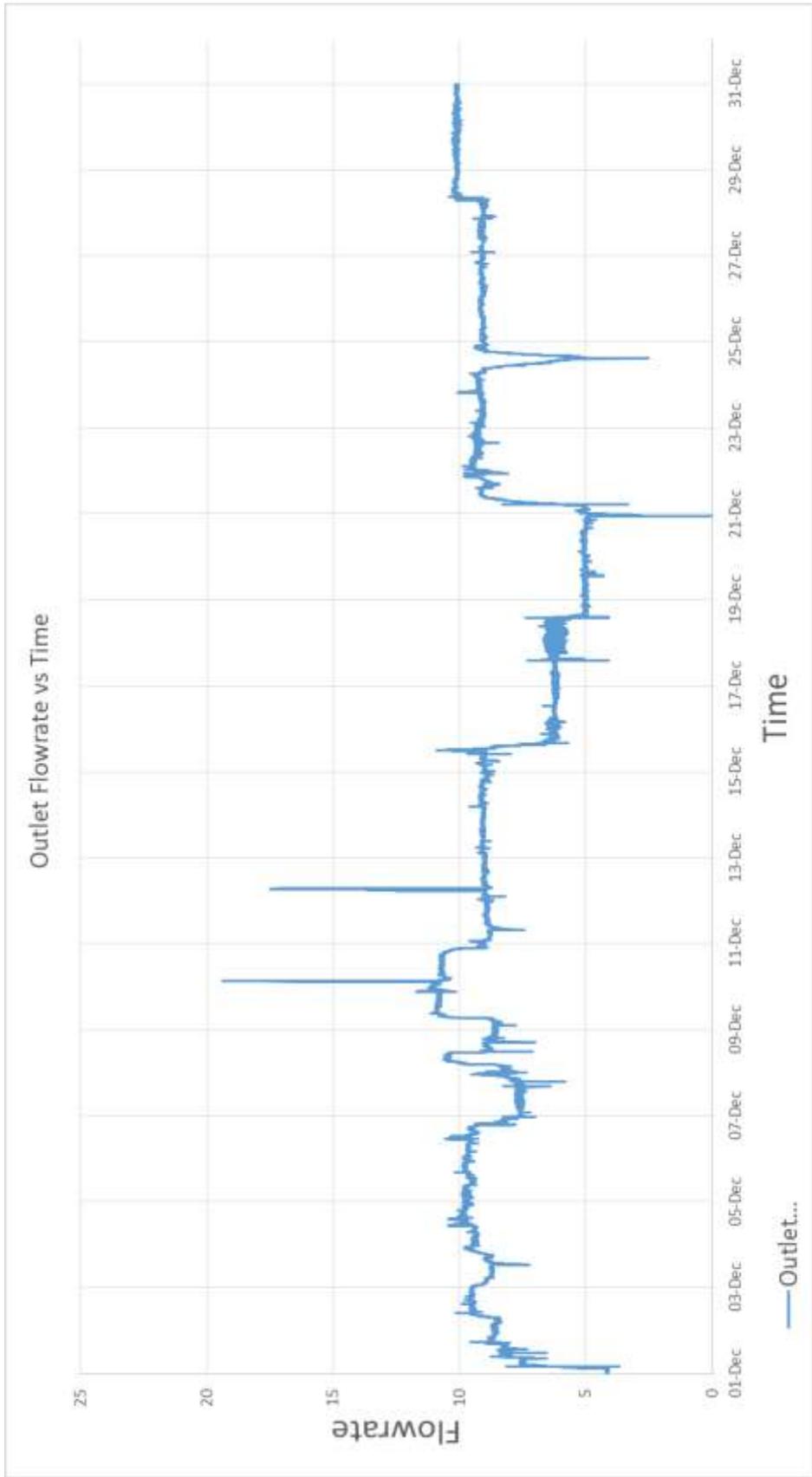

Figure 4.2 Graph of Outlet Flowrate against Time

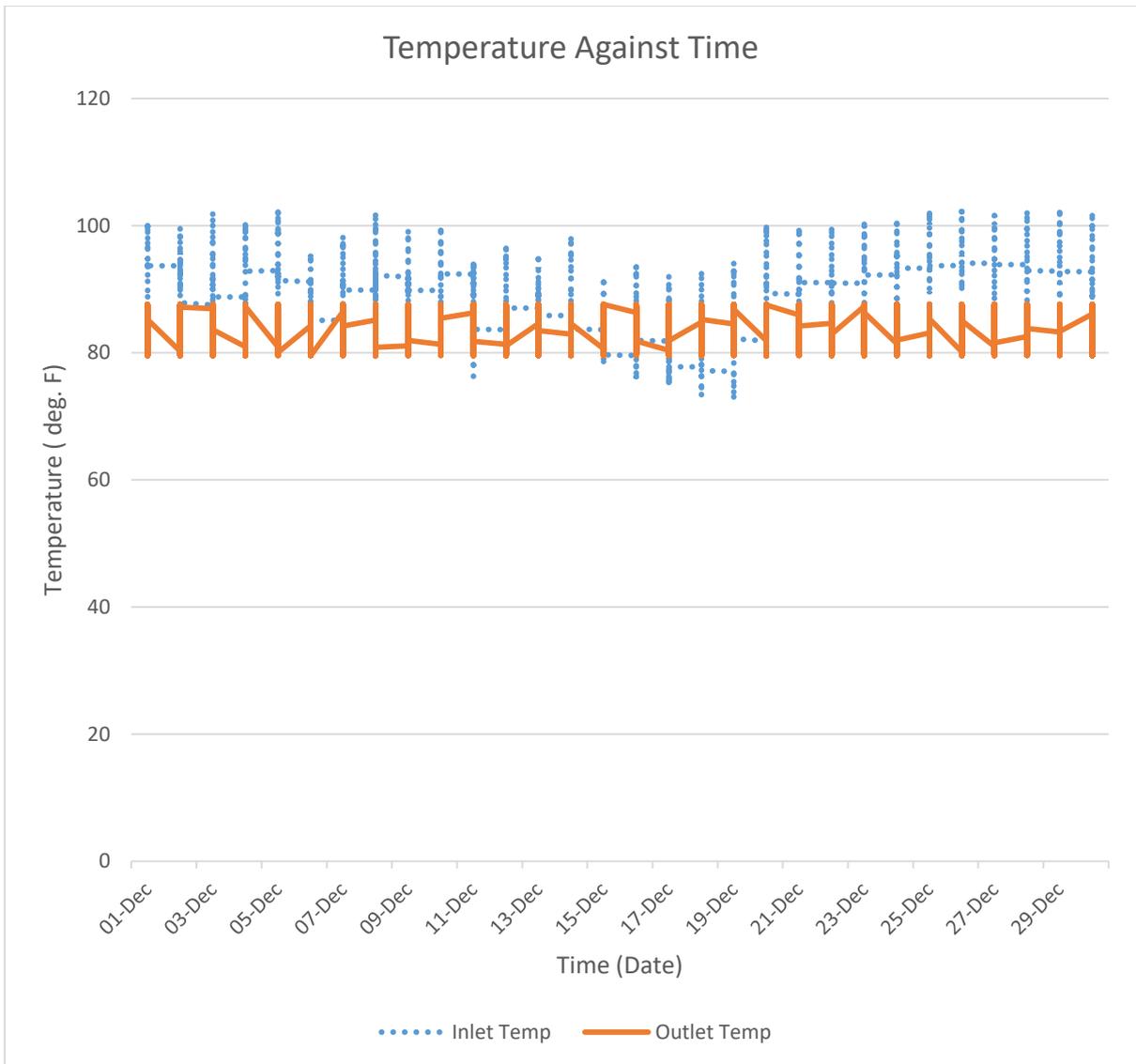

Figure 4.3 Graph of Average Daily Inlet and Outlet Temperature against Time

### 4.1.2 Data Validation

Figure 4.4 showed that there was a close fit between measured flowrate and predicted flowrate with a coefficient of correlation score of 0.95 despite limited knowledge about the pipeline parameters. This validates the measured pipeline operational parameters and ensures that the analysis is done on competent data. Table 4.1 shows the descriptive statistics of the pipeline used for carrying out this analysis. As expected, the standard deviation and the range of the inlet pressure for the time period is much less than that of the outlet pressure. The range and standard deviation of the inlet temperature are much higher than that of the outlet temperature. It must be noted that the compressibility is almost always constant with a low standard deviation from the mean value of 0.72.

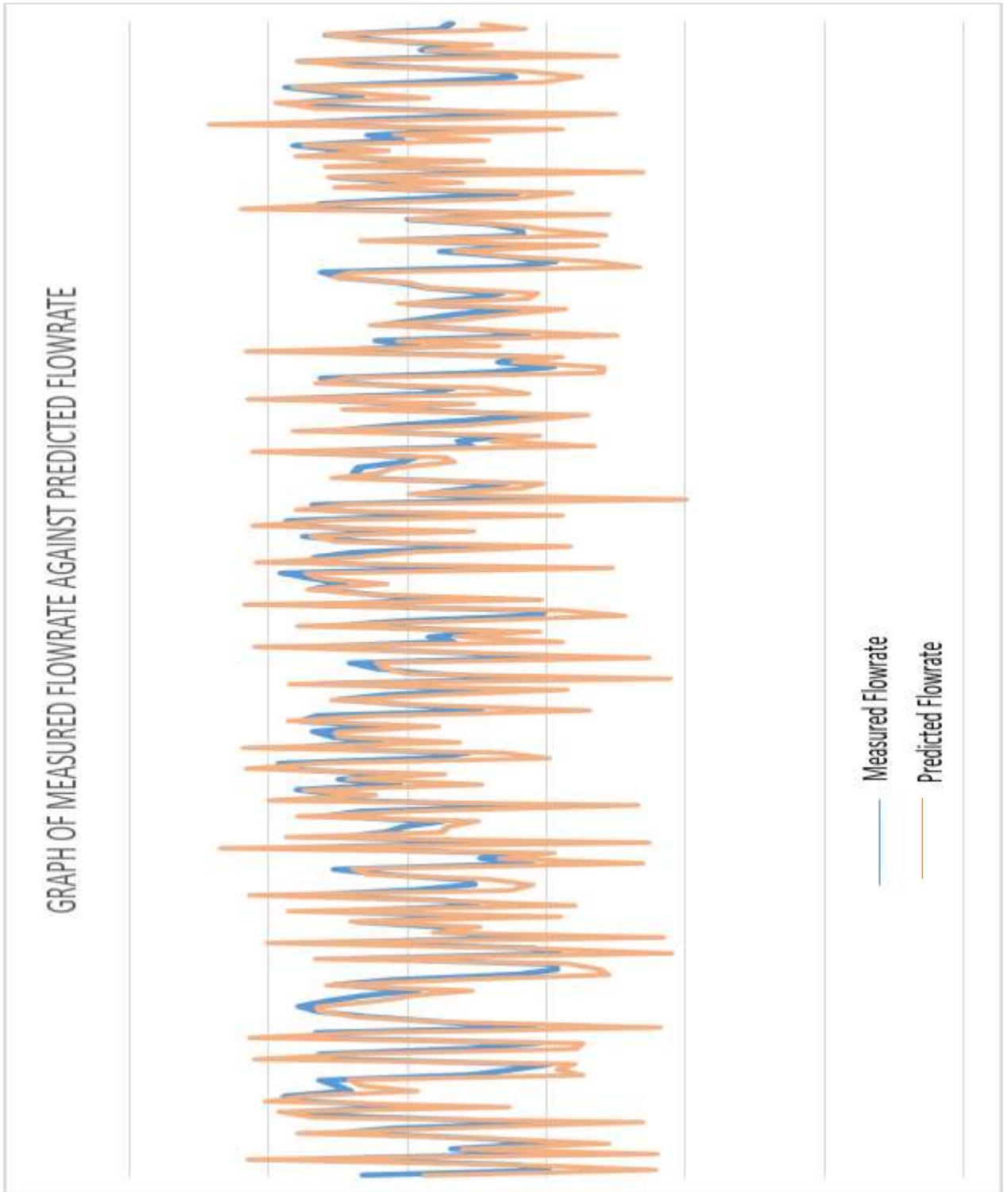

Figure 4.4 Fit of Measured Flowrate on Predicted Flowrate By Transient Simulator (Zoomed).

Table 4.1: Table of Descriptive Properties for the Test Pipeline

|  | Inlet Pressure (psia) | Inlet Temp (deg F) | Z | Viscosity (cp) | Outlet Pressure (psia) | Outlet Temp (deg F) | Outlet Flow-rate (mmscm) | SG | Inlet Flow-rate (mmscm) |
|---|---|---|---|---|---|---|---|---|---|
| mean | 1317.60 | 90.71 | 0.72 | 0.02 | 1269.96 | 83.56 | 8.54 | 0.63 | 13.20 |
| std dev | 11.40 | 6.25 | 0.01 | 0.00 | 24.16 | 2.35 | 1.64 | 0.01 | 3.26 |
| min | 1256.50 | 73.02 | 0.68 | 0.02 | 1075.97 | 79.52 | 2.97 | 0.61 | 3.87 |
| 25% | 1309.79 | 87.13 | 0.71 | 0.02 | 1254.77 | 81.52 | 8.02 | 0.62 | 11.08 |
| median | 1315.48 | 91.02 | 0.72 | 0.02 | 1265.52 | 83.55 | 9.06 | 0.63 | 13.33 |
| 75% | 1325.43 | 95.03 | 0.73 | 0.02 | 1284.22 | 85.61 | 9.47 | 0.63 | 15.64 |
| max | 1366.67 | 102.54 | 0.74 | 0.02 | 1352.11 | 87.62 | 18.90 | 0.66 | 32.05 |

### 4.1.3 Model Validation

Table 4.2 shows the statistical metrics for the model validation. It can be seen the artificial neural network fitted the data best followed by the support vector machine. This is consistent with the theoretical belief that these two models are function approximators and given the right architecture, they can fit any data set perfectly. However, because of their internal complexities, they may not necessarily provide the best results as regards leak detection. It can be seen that random forest and decision tree are the most sensitive as they detect a leak of 0.1% of nominal in about 2 hours while the others could not detect the leak despite having over four hours of leak data to analyse as shown in Figure 4.5 & Figure 4.6.

In terms of reliability, since the models were tuned to ensure no false alarms, they can all be said to be reliable. Once there is an alarm, there is an almost 100% chance that there is a leak. Since the models act as a kind of volume balance system, accuracy can only be measured in terms of estimating leak flow rate as a volume balance system cannot localize leaks. It can be seen that generally, the accuracy of the models are low and they tend to improve as the leak gets closer to the outlet. This result is consistent with general LDS performance. The models have sacrificed some accuracy in order to boost reliability. Since only inlet and outlet parameters are used with no knowledge of valve behaviour, compressor action etc., robustness can be said to be at least good since the analysis was carried out on real field data.

Table 4.3 shows the average time to leak detection for each leak size in comparison with a literature data. The literature data reports time to detection for three different types of typical pipeline systems at different flow rates and thus an average time for similar leak flowrate ratio should give an objective benchmark for comparison. Similarly, the literature data simulated close to real field data taking into account phenomenon likes equipment drift, noise etc. While the sampling time of the literature data is different from that of the intelligent models, theoretically, this should affect little in terms of time to leak detection while it should have more effect on accuracy. It should be noted that the real time transient model found in literature generally has a better localization precision than the intelligent models as shown in Table 4.4. This is likely because the localization formula was a linearized approximation. Table 4.5 shows the rankings of the intelligent models in terms of the Sensitivity-Accuracy-Reliability-Robustness (SARR) trade-offs. Generally, intelligent models are more robust and reliable while RTTM are more sensitive and accurate. The biggest trade-off seems to be the accuracy-reliability trade-off. It is also useful to note that tree-based ensemble models seem to be as adept to leak detection as complex architectures such as SVM and neural networks. In conclusion, it can be seen that the intelligent models generally performance well in terms of leak detection and have almost similar performance to the real time transient model found in data.

Table 4.2 Table of Models' Accuracy Using Statistical Metrics

|  | GB | RF | DT | ANN | SVM |
|---|---|---|---|---|---|
| **RMSE** | 0.097 | 0.106 | 0.138 | 0.072 | 0.083 |
| **MAE** | 0.053 | 0.038 | 0.052 | 0.034 | 0.048 |
| **$R^2$ Score (train)** | 0.999 | 0.999 | 1.000 | 0.997 | 0.997 |
| **$R^2$ Score (Test)** | 0.997 | 0.995 | 0.993 | 0.998 | 0.998 |
| **$R^2$ Score (CV)** | 0.994 | 0.996 | 0.991 | 0.997 | 0.996 |

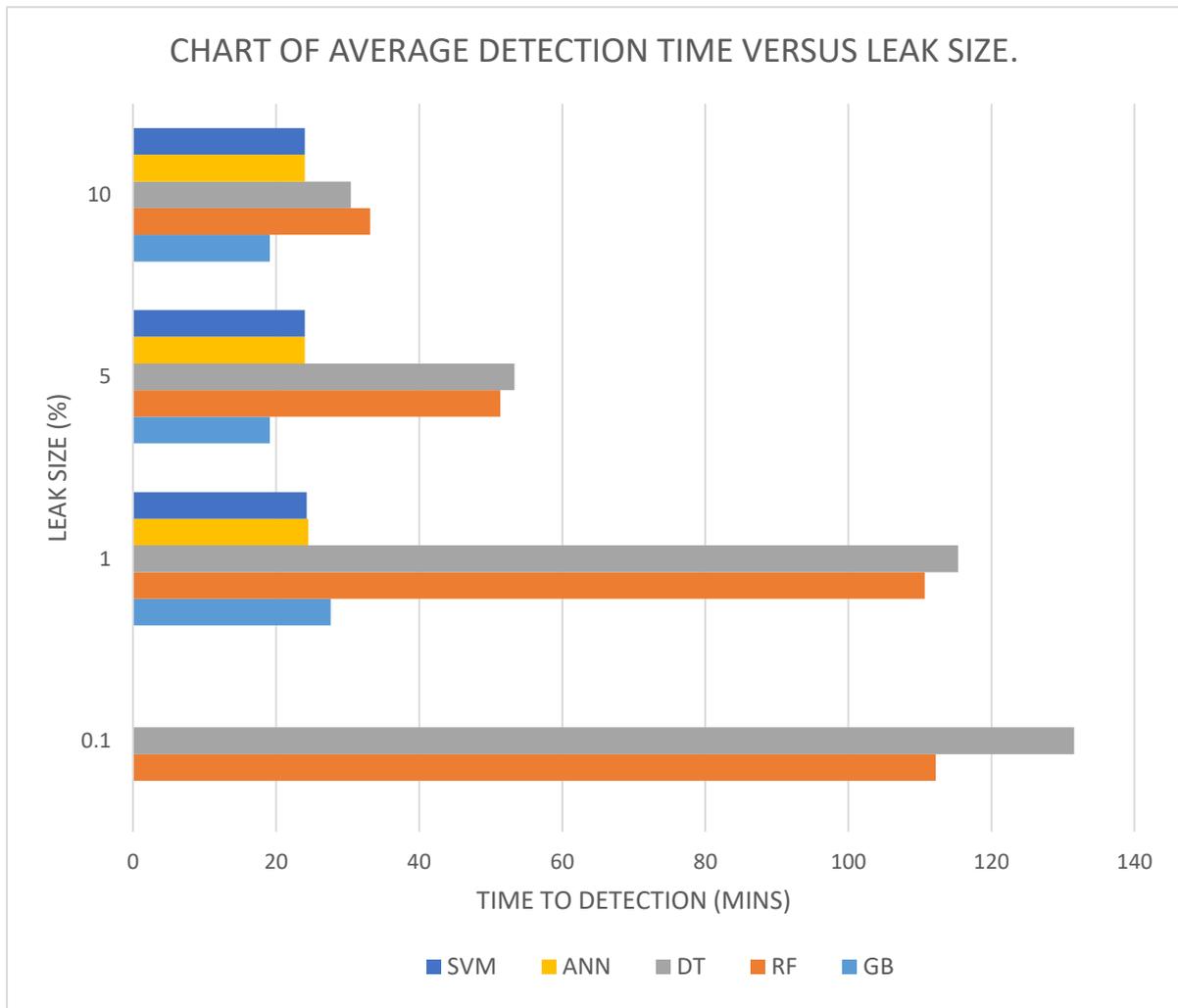

Figure 4.5 Figure of Average Detection Time Versus Leak Size For Each Model.

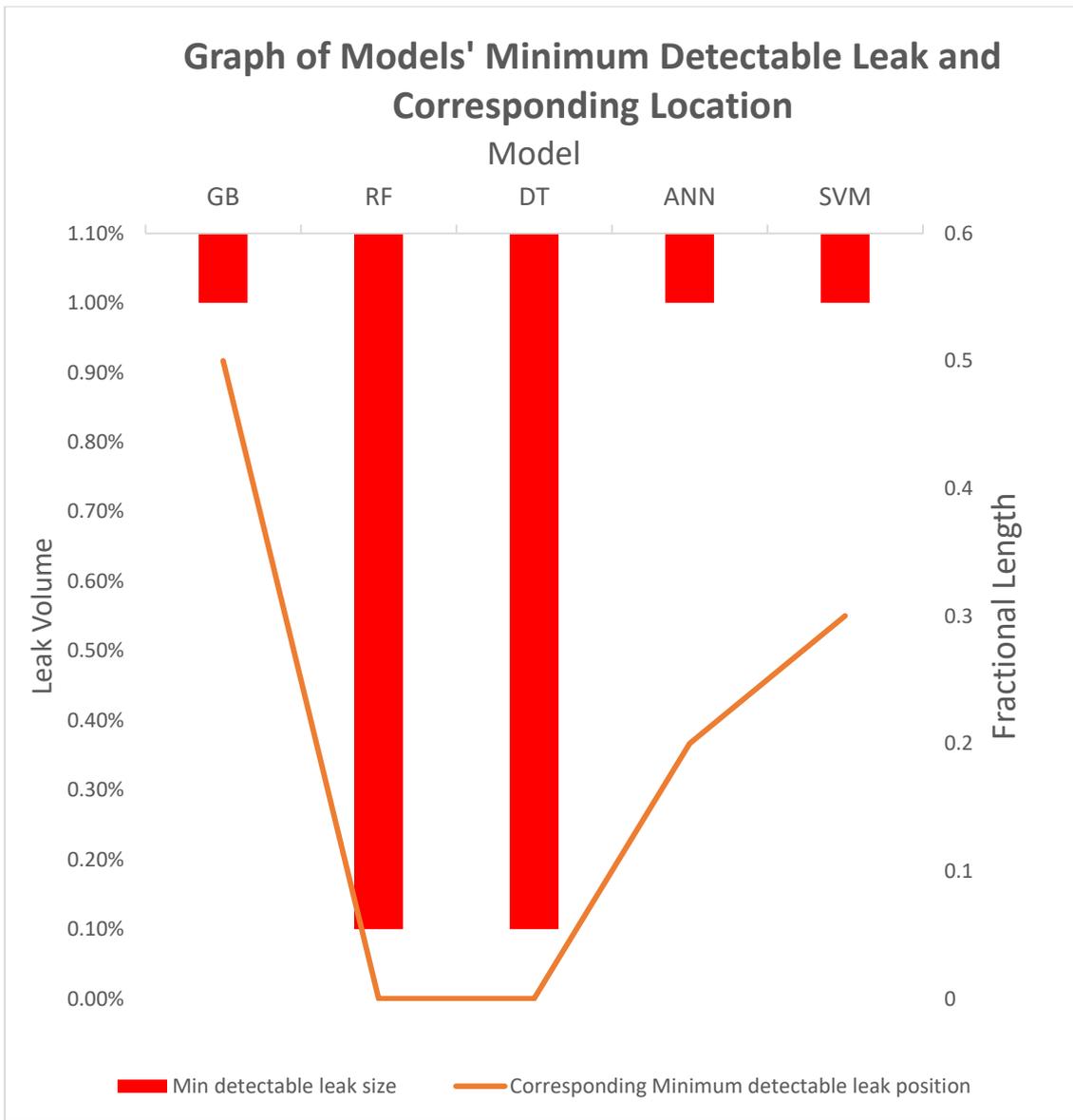

Figure 4.6 Graph Showing Models' Minimum Detectable Leak and Corresponding Minimum Detectable Location

Table 4.3 Table of Average Time to Detection in Minutes for Each Leak Size per Model.

| Leak Size (%) | GB | RF | DT | ANN | SVM | RTTM |
|---|---|---|---|---|---|---|
| 0.1 | N/A* | 112 | 132 | N/A* | N/A* | 173 |
| 1 | 28 | 111 | 115 | 25 | 24 | 100 |
| 5 | 19 | 51 | 53 | 24 | 24 | 36 |
| 10 | 19 | 33 | 30 | 24 | 24 | 32 |

* Model could not detect leak in four hours of leak signature

Table 4.4 Table of Average Leak Localization Error for Each Model.

| Model/ Leak Size | Gradient Boosting | Random Forest | Decision Trees | Support Vector Machines | Perceptron Neural Network | RTTM |
|---|---|---|---|---|---|---|
| 0.10% | n/a | n/a | n/a | n/a | n/a | n/a |
| 1% | n/a | 21% | 15% | 17% | 23% | 10% |
| 5% | 14% | 18% | 12% | 15% | 18% | 8% |
| 10% | 12% | 13% | 10% | 12% | 15% | 7% |

Table 4.5 Table Showing the SARR Rankings of Each Model.

| | Sensitivity | Accuracy | Robustness | Reliability | Ranking** |
|---|---|---|---|---|---|
| **Random Forest** | 2 | 5 | 1 | 4 | 12 |
| **Real Time Transient Model** | 1 | 1 | 5 | 6 | 13 |
| **Gradient Boosting** | 6 | 3 | 1 | 3 | 13 |
| **Perceptron Neural Network** | 4 | 6 | 3 | 1 | 14 |
| **Support Vector Machines** | 5 | 4 | 3 | 2 | 14 |
| **Decision Tree** | 3 | 2 | 5 | 5 | 15 |

** Lower is better.

# CHAPTER FIVE

## CONCLUSION AND RECOMMENDATION

### 5.1 CONCLUSION

Big data analytics is gaining more and more relevance in the energy industry today as the Oil and Gas industry seeks to make better use of the terabytes of data it generates on a daily basis. This project aimed at using data analytics alongside intelligent models to solve the very important and persistent problem of leak detection in natural gas pipelines. The intelligent models used basic pipeline operational parameters alongside a simple volume balance leak detection system to detect a leak and determine its volume in a two-stage regresso-classification hierarchical model where the intelligent model acts as a regressor and the modified logistic regression acts as a classifier.

The results obtained were very promising. The intelligent models performed relatively well when compared to a real-time transient model (which is currently the industry preferred solution for leak detection in natural gas pipelines) obtained in literature. Recently, researchers have sought to merge the leak detection methodology of statistical analysis and real-time transient analysis to form a field of Extended Real-Time Transient Model (E-RTTM) in order to improve leak detection results. This research suggests intelligent models which are easier to model than pure statistical models could be used alongside RTTM to improve leak detection results.

### 5.2 RECOMMENDATION

The results show that more work can be done to investigate the usefulness of intelligent models in leak detection in natural gas pipelines with bigger data volumes and much complex networks. Similarly, the result obtained suggests that intelligent models could be used alongside real-time transient model in an E-RTTM in order to significantly improve leak detection results. Lastly, since intelligent models are data-driven, the application of intelligent models in new pipelines with little or no operational data should also be studied. It is therefore recommended that more research can be done in these areas.

63 | P a g e# REFERENCES

Afebu K.O., Abbas A.J, Nasr G.G. & Kadir, 2015 A. Integrated Leak Detection in Gas Pipelines Using OLGA Simulator and Artificial Neural Network. Paper was prepared to be presented at the Abu Dhabi International Petroleum Exhibition and Conference held in Abu Dhabi, UAE, 9 – 12 November 2015.

Ahmad A., Abd. Hamid M., 2003 Pipeline Leak Detection System in a Palm Oil Fractionation Plant Using Artificial Neural Network.

Alaska Department of Environmental Conservation, n.d. *Technical Review of Leak Detection Technologies,* s.l.: Alaska Department of Environmental Conservation.

Anon. Boosting Models [Online].
Available at: https://www.jianshu.com/p/7df05ef1ef59

Anon, 2017.*Inductive Automation.* [Online]
Available at: https://inductiveautomation.com/what-is-scada [Accessed 10 May 2017].

Anon, 2018. Neuro-fuzzy Survey [Online].
Available at: https://studylib.net/doc/18324186/neuro-fuzzy-systems--a-survey#

Balda Rivas, K. V., Civan, F. & Compnies, W., 2013. Application of Mass Balance and Transient FLow Modeling for Leak Detection in Liquid Pipelines, Oklahoma: Society of Petroleum Enginners, SPE 164520.

Belsito S., Lommbardi P., Andreussi P., and Banerjee S. 1998 Leak Detection in Liquefied Gas Pipeline by Artificial Neural Networks AIChE Journal Vol 44, Issue 12

Bishop C.M. and Tipping M.E, 2016. Bayesian Regression and Classification [Online]
Available at: https://www.microsoft.com/en-us/research/wp-content/uploads/2016/02/bishop-nato-bayes.pdf

# APPENDIX

Python code for implementing the leak detection models described in chapter 3.

```python
# coding: utf-8

import pandas as pd
import numpy as np
import mglearn

data=pd.read_csv(r"C:\Users\ADEBAYO\Desktop\Project\Project excelsheets\most recent\projectdata\Field 2.csv")
leak=pd.read_csv(r"C:\Users\ADEBAYO\Desktop\Project\Project excelsheets\most recent\projectdata\Leak_.csv")

import seaborn as sns
sns.pairplot(data)

data.describe()
print("Original features:\n", list(data.columns), "\n")

feature_col_names=['Inlet Pressure', 'Inlet Temp', 'Outlet Pressure', 'Outlet Temp']
predicted_class_names=['Flowrate']

X_train=data[feature_col_names].values
Y_train=data[predicted_class_names].values

Y_train=Y_train.ravel()

split_test_size=0.30

from sklearn.model_selection import train_test_split
Xtrain, Xtest, Ytrain, Ytest= train_test_split(X_train,Y_train, test_size=split_test_size, random_state=12)

from sklearn.ensemble import GradientBoostingRegressor,RandomForestRegressor
from sklearn.tree import DecisionTreeRegressor
from sklearn.svm import SVR
from sklearn.neural_network import MLPRegressor

# # LEAK SIMULATION

def leaksimul(Lld=0,Qld=0):
    if (Qld==0 or Lld==0):
        leak['Outlet Pressure'][:]=leak['P2'][:]
    else:
        Fl=round((1+(Lld*(Qld**2+Qld*2)))**-0.5,4)
        leak['Outlet Pressure'][30:]=(leak['Inlet Pressure'][30:]**2-(leak['Flowrate'][30:]/(0.0244*Fl))**2)**0.5
        leak['Outlet Pressure'][0:30]=leak['P2'][0:30]
    return
```

## # LEAK DETECTION

```python
from math import exp as exp
def leak_detect(observed=None,predicted=None):
    leak_index=0
    leak_profile=[0,0]
    b=0
    for index in range(20,len(leak)):
        leak['leak index']=(abs(Yleak-Ypred)>(MAE+0.01)).ravel()
        if (leak['leak index'][index]==True):
            leak_sub=leak[index-20:index]
            a=(leak_sub['leak index']==True).sum() + 1
            leak_ind=exp(-1*(1/(1+a**2)))
            #print (leak_profile)
            if (leak_ind>leak_index):
                leak_index=round(leak_ind,2)
                leak_profile = [leak_index,index]
            if (leak_index>0.99):
                b=b+1
            else:
                b=0
            if (b>3):
                print('LEAK')
                i=leak_profile[1]
                leak_details=[leak_index,round(abs(Yleak[i]-Ypred[i])*100/Yleak[i],2),round(leak['Inlet Pressure'][i],2),
                              round(leak['Outlet Pressure'][i],2),(i-32)*2]
                print('leak index, percentage leak, Inlet Pressure, Outlet Pressure, minutes to detect:', leak_details)
                #print("% leak is: ",leak_details[1])
                #print("leak inlet pressure is: ",leak_details[2])
                #print("leak outlet pressure is: ",leak_details[3])
                #print(Yleak[i])
                break

    print('end of simulation')

    return

from sklearn.metrics import mean_squared_error,mean_absolute_error
from sklearn.model_selection import cross_val_score,KFold,StratifiedKFold
kfold=KFold(n_splits=5)
from sklearn.preprocessing import PolynomialFeatures,MinMaxScaler
poly=PolynomialFeatures(degree=2).fit(Xtrain)
scaler=MinMaxScaler()
```

## # ## Gradient Boosting

# #### TRAINING

```
import mglearn
Utrain=(poly.transform(Xtrain))
Utest=(poly.transform(Xtest))
gbrt = GradientBoostingRegressor()
from sklearn.model_selection import GridSearchCV
gs=GridSearchCV(gbrt,param_grid,cv=kfold).fit(Utrain,Ytrain)
Ypred=gs.predict(Utest)
MAE=mean_absolute_error(Ytest,Ypred)
MSE=mean_squared_error(Ytest,Ypred)
print("GBR:", np.sqrt(MSE),MAE)
print("GBR Training set score: {:.5f}".format(gs.score(Utrain,Ytrain)))
print("GBR Test set score: {:.5f}".format(gs.score(Utest,Ytest)))
results=pd.DataFrame(gs.cv_results_)
scores=np.array(results.mean_test_score).reshape(4,3)
mglearn.tools.heatmap(scores,xlabel='max_features',xticklabels=param_grid['max_features']
            ,ylabel='n_estimators',yticklabels=param_grid['n_estimators'],
            cmap='viridis',fmt='%0.5f')
```

# #### Leak Prediction

```
for i in [0.001,0.01,0.05,0.1]:
    for j in [0.1,0.2,.3,.4,.5,.6,.7,.8,.9]:
        leaksimul(Qld=i,Lld=j)
        Xleak=poly.transform(leak[feature_col_names].values);
        Yleak=(leak[predicted_class_names].values).ravel();
        Ypred=(gbrt.predict(Xleak));
        leak_detect(observed=Yleak,predicted=Ypred)
        print ('This is Qld and Lld:',i,j)
```

# # Random Forest

# ### Training

```
param_grid = {'n_estimators':[50,200,350,500],
        'learning_rate':[.1,1,10]}
Utrain=(poly.transform(Xtrain))
Utest=(poly.transform(Xtest))
rf = RandomForestRegressor()
gs=GridSearchCV(rf,param_grid,cv=kfold).fit(Utrain,Ytrain)
Ypred=gs.predict(Utest)
MAE=mean_absolute_error(Ytest,Ypred)
MSE=mean_squared_error(Ytest,Ypred)
print("GBR:", np.sqrt(MSE),MAE)
print("GBR Training set score: {:.5f}".format(gs.score(Utrain,Ytrain)))
print("GBR Test set score: {:.5f}".format(gs.score(Utest,Ytest)))
```

```python
results=pd.DataFrame(gs.cv_results_)
scores=np.array(results.mean_test_score).reshape(3,3)
mglearn.tools.heatmap(scores,xlabel='max_features',xticklabels=param_grid['max_features']
                      ,ylabel='n_estimators',yticklabels=param_grid['n_estimators'],
                      cmap='viridis',fmt='%0.5f')
```

# ### Leak detection

```python
for i in [0.001,0.01,0.05,0.1]:
    for j in [0.1,0.2,.3,.4,.5,.6,.7,.8,.9]:
        leaksimul(Qld=i,Lld=j)
        Xleak=poly.transform(leak[feature_col_names].values);
        Yleak=(leak[predicted_class_names].values).ravel();
        Ypred=(gs.predict(Xleak));
        leak_detect(observed=Yleak,predicted=Ypred)
        print ('This is Qld and Lld:',i,j)
```

# ## DECISION TREE

# ### Training

```python
param_grid = {'max_features':[None,'log2','auto'],
              'min_samples_split':[2,5,10]}
Utrain=(poly.transform(Xtrain))
Utest=(poly.transform(Xtest))
tree = DecisionTreeRegressor()
from sklearn.model_selection import GridSearchCV
gs=GridSearchCV(tree,param_grid,cv=kfold).fit(Utrain,Ytrain)
Ypred=gs.predict(Utest)
MAE=mean_absolute_error(Ytest,Ypred)
MSE=mean_squared_error(Ytest,Ypred)
print("GBR:", np.sqrt(MSE),MAE)
print("GBR Training set score: {:.5f}".format(gs.score(Utrain,Ytrain)))
print("GBR Test set score: {:.5f}".format(gs.score(Utest,Ytest)))
results=pd.DataFrame(gs.cv_results_)
scores=np.array(results.mean_test_score).reshape(3,3)
mglearn.tools.heatmap(scores,xlabel='max_features',xticklabels=param_grid['max_features']
                      ,ylabel='min_samples_split',
                       yticklabels=param_grid['min_samples_split'],
                      cmap='viridis',fmt='%0.5f')
```

# ### Leak Detection

```python
for i in [0.001,0.01,0.05,0.1]:
    for j in [0.1,0.2,.3,.4,.5,.6,.7,.8,.9]:
        leaksimul(Qld=i,Lld=j)
        Xleak=poly.transform(leak[feature_col_names].values);
        Yleak=(leak[predicted_class_names].values).ravel();
```

```
        Ypred=(gs.predict(Xleak));
        leak_detect(observed=Yleak,predicted=Ypred)
print ('This is Qld and Lld:',i,j)

# ## SUPPORT VECTOR MACHINE

# ### Training

from sklearn.preprocessing import MinMaxScaler
sc=MinMaxScaler().fit(poly.transform(Xtrain))
Utrain=sc.transform(poly.transform(Xtrain))
Utest=sc.transform(poly.transform(Xtest))

param_grid = {'C':[.1,1,10,100,1000],
        'kernel':['rbf','linear']}
Utrain=(poly.transform(Xtrain))
Utest=(poly.transform(Xtest))
svm = SVR(gamma='scale')
from sklearn.model_selection import GridSearchCV
gs=GridSearchCV(svm,param_grid,cv=kfold).fit(Utrain,Ytrain)
Ypred=gs.predict(Utest)
MAE=mean_absolute_error(Ytest,Ypred)
MSE=mean_squared_error(Ytest,Ypred)
print("GBR:", np.sqrt(MSE),MAE)
print("GBR Training set score: {:.5f}".format(gs.score(Utrain,Ytrain)))
print("GBR Test set score: {:.5f}".format(gs.score(Utest,Ytest)))
results=pd.DataFrame(gs.cv_results_)

scores=np.array(results.mean_test_score).reshape(1,5)
mglearn.tools.heatmap(scores,xlabel='C',xticklabels=param_grid['C']
            ,ylabel='kernel',yticklabels=param_grid['kernel'],
            cmap='viridis',fmt='%0.5f')

# ### Leak Detection

for i in [0.001,0.01,0.05,0.1]:
    for j in [0.1,0.2,.3,.4,.5,.6,.7,.8,.9]:
        leaksimul(Qld=i,Lld=j)
        Xleak=sc.transform(poly.transform(leak[feature_col_names].values));
        Yleak=(leak[predicted_class_names].values).ravel();
        Ypred=(gs.predict(Xleak));
        leak_detect(observed=Yleak,predicted=Ypred)
        print ('This is Qld and Lld:',i,j)

# ## NEURAL NETWORK

# ### MLP
```

```python
from sklearn.preprocessing import MinMaxScaler
sc=MinMaxScaler().fit((Xtrain))
Utrain=sc.transform((Xtrain))
Utest=sc.transform((Xtest))

param_grid = {'hidden_layer_sizes':[(5,),(10,),(20,)
                      ],
       'alpha':[0.01,1,10]}
mlp = MLPRegressor(max_iter=1000,momentum=0.1,solver='lbfgs')
from sklearn.model_selection import GridSearchCV
gs=GridSearchCV(mlp,param_grid,cv=kfold).fit(Utrain,Ytrain)
Ypred=gs.predict(Utest)
MAE=mean_absolute_error(Ytest,Ypred)
MSE=mean_squared_error(Ytest,Ypred)
print("GBR:", np.sqrt(MSE),MAE)
print("GBR Training set score: {:.5f}".format(gs.score(Utrain,Ytrain)))
print("GBR Test set score: {:.5f}".format(gs.score(Utest,Ytest)))
results=pd.DataFrame(gs.cv_results_)
scores=np.array(results.mean_test_score).reshape(3,3)
mglearn.tools.heatmap(scores,xlabel='hidden_layer_sizes',
            xticklabels=param_grid['hidden_layer_sizes']
            ,ylabel='alpha',yticklabels=param_grid['alpha'],
            cmap='viridis',fmt='%0.5f')

for i in [0.001,0.01,0.05,0.1]:
   for j in [0.1,0.2,.3,.4,.5,.6,.7,.8,.9]:
      leaksimul(Qld=i,Lld=j)
      Xleak=sc.transform(leak[feature_col_names].values);
      Yleak=(leak[predicted_class_names].values).ravel();
      Ypred=(gs.predict(Xleak));
      leak_detect(observed=Yleak,predicted=Ypred)
      print ('This is Qld and Lld:',I,j)
```